\documentclass[lettersize,journal]{IEEEtran}
\usepackage{amsmath,amsfonts}
\usepackage{algorithmic}
\usepackage{algorithm}
\usepackage{array}
\usepackage{textcomp}
\usepackage{stfloats}
\usepackage{url}
\usepackage{verbatim}
\usepackage{graphicx}
\usepackage{cite}

\usepackage{times}
\usepackage{amssymb}
\usepackage{booktabs}
\usepackage{multirow}
\usepackage{pifont}
\usepackage{subcaption}
\usepackage{color, colortbl}
\usepackage{xcolor}
\usepackage{caption}

\definecolor{Grey}{gray}{0.6}
\definecolor{lavender(web)}{rgb}{0.9, 0.9, 0.98}
\definecolor{beige}{rgb}{0.98, 0.98, 0.88}%{0.96, 0.9, 0.83}
\def\method{VIPA}

\newcommand{\cmark}{\ding{51}}
\newcommand{\xmark}{\ding{55}}

 \newcolumntype{L}[1]{>{\raggedright\let\newline\\\arraybackslash\hspace{0pt}}m{#1}}
 \newcolumntype{C}[1]{>{\centering\let\newline\\\arraybackslash\hspace{0pt}}m{#1}}
 \newcolumntype{R}[1]{>{\raggedleft\let\newline\\\arraybackslash\hspace{0pt}}m{#1}}
 
\begin{document}

\title{VIPA: Visual Informative Part Attention for Referring Image Segmentation}

% \author{{}
\author{Yubin Cho\thanks{Yubin Cho and Hyunwoo Yu contributed equally to this work. (Corresponding author: Suk-Ju Kang.)}, Hyunwoo Yu, Kyeongbo Kong, Kyomin Sohn, Bongjoon Hyun and Suk-Ju Kang,~\IEEEmembership{Member,~IEEE}
% \author{ }
        % <-this % stops a space
\thanks{Yubin Cho was with the School of Artificial Intelligence, Sogang University (e-mail: dbqls1219@sogang.ac.kr) and she is currently with the AI Lab of CTO division, LG Electronics (e-mail: ubin.cho@lge.com).}
\thanks{
Hyunwoo Yu, and Suk-Ju Kang are with the School of Electronic Engineering, Sogang University (e-mail: hyunwoo137@sogang.ac.kr; sjkang@sogang.ac.kr).}% <-this % stops a space
\thanks{
Kyeongbo Kong is with the School of Electronic Engineering, Pusan University (e-mail: kbkong@pusan.ac.kr).}
\thanks{
Kyomin Sohn and Bongjoon Hyun are with the Memory Business Division, Samsung Electronics (e-mail: kyomin.sohn@samsung.com; bongj.hyun@samsung.com).}

}

\markboth{Journal of \LaTeX\ Class Files,~Vol.~14, No.~8, August~2021}%
{CHO \MakeLowercase{\textit{et al.}}: Visual Informative Part Attention for Referring Image Segmentation}


\maketitle

\begin{abstract}
Referring Image Segmentation (RIS) aims to segment a target object described by a natural language expression. Existing methods have evolved by leveraging the vision information into the language tokens. To more effectively exploit visual contexts for fine-grained segmentation, we propose a novel Visual Informative Part Attention (VIPA) framework for referring image segmentation. VIPA leverages the informative parts of visual contexts, called a visual expression, which can effectively provide the structural and semantic visual target information to the network. This design reduces high-variance cross-modal projection and enhances semantic consistency in an attention mechanism of the referring image segmentation. We also design a visual expression generator (VEG) module, which retrieves informative visual tokens via local-global linguistic context cues and refines the retrieved tokens for reducing noise information and sharing informative visual attributes. This module allows the visual expression to consider comprehensive contexts and capture semantic visual contexts of informative regions. In this way, our framework enables the network's attention to robustly align with the fine-grained regions of interest. Extensive experiments and visual analysis demonstrate the effectiveness of our approach. Our VIPA outperforms the existing state-of-the-art methods on four public RIS benchmarks.
\end{abstract}


\section{Introduction} \label{sec:intro}

\begin{figure*}[t]
\includegraphics[width=\linewidth]{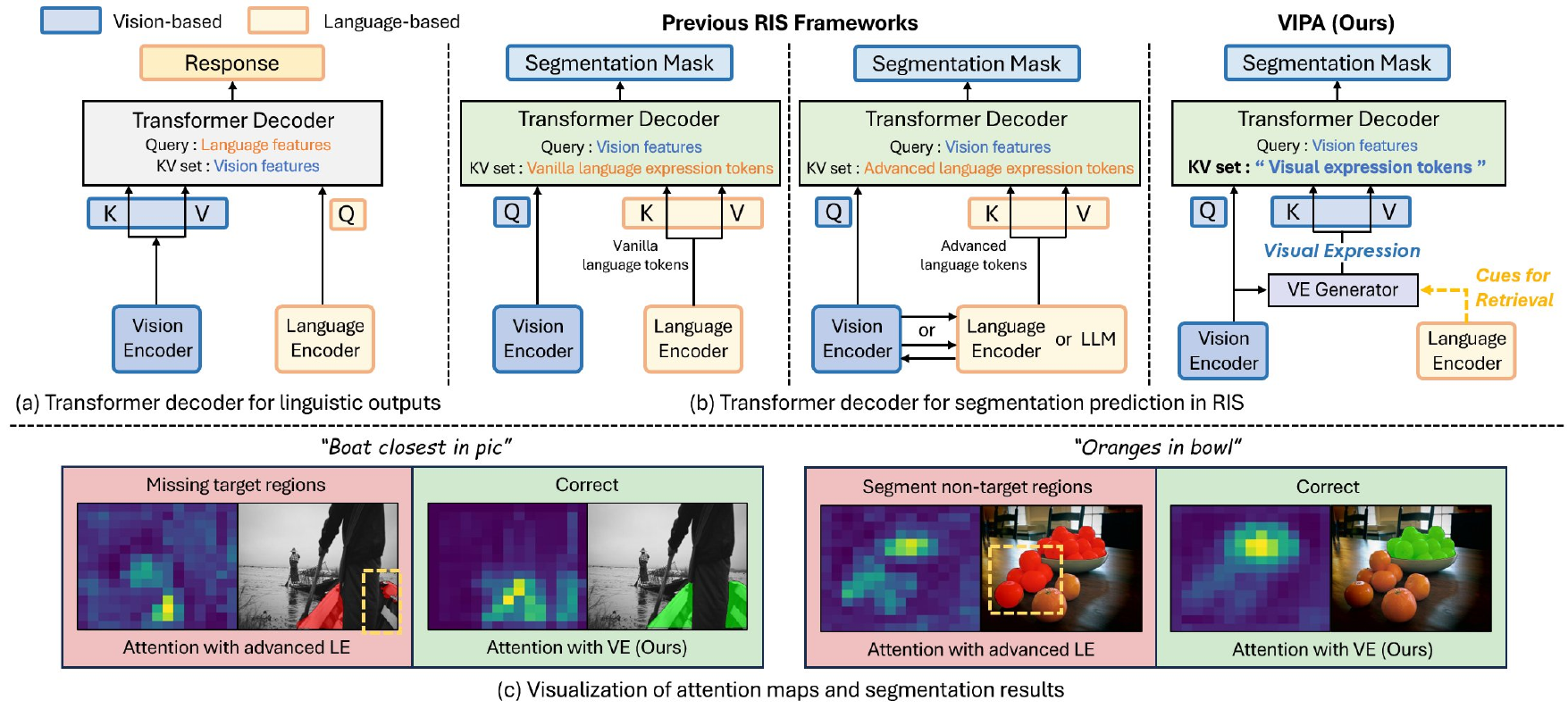} 
\caption{(a) Example of a transformer decoder used to extract language outputs. (b) Illustration of different RIS frameworks. Different from previous works, our approach leverages \textit{visual expression}, generated from the retrieved informative parts of visual contexts, as a key-value set in the Transformer-based segmentation decoder. (c) Visual comparison of two different key-value sets. Yellow dotted boxes are incorrect predictions. The visual expression (VE) robustly guides the network's attention to the regions of interest by enhancing semantic coherence in the attention mechanism of referring image segmentation, whereas the advanced language expression (LE) results in incomplete predictions.}
\label{_intro}
\end{figure*}

Referring image segmentation (RIS) \cite{liu2021cross, hu2016segmentation, liu2017recurrent} is one of the challenging vision-language tasks and can be applied in various applications such as human-robot interaction and the object retrieval. Given an image and a natural language expression describing a target object within the image, the key point in this task is for the network to precisely segment the target object regions from the image by referencing the given target information. 

As shown in Figs. \ref{_intro} (a) and (b), the role of a Transformer decoder differs for a certain purpose. Previous RIS methods \cite{wang2022cris, hu2023beyond, lai2024lisa} have adopted a Transformer-based segmentation decoder using vision features as queries and language tokens as keys-values to find the target regions by referencing the given language information. As a key-value set, some studies \cite{ding2022vlt, hu2023beyond, chng2024mask, wu2024toward, li2025bring, xu2024cmirnet, shang2022cross, xiao2025temporal, ding2023bilateral} use the advanced language tokens extracted by the bidirectional cross-attention encoders to enhance alignment with cross-modal information. More recent studies \cite{lai2024lisa, ren2024pixellm, xia2024gsva, rasheed2024glamm} employ LLMs \cite{touvron2023llama2, chiang2023vicuna} to improve understanding of the language expression via LLM’s immense knowledge, and exploit the generated language tokens in the segmentation decoder, as shown in Fig. \ref{_intro}(b).

In this task, we found that how to construct the key-value information when the query is the vision feature in the segmentation decoder has a significant impact on segmentation performance. Based on this perspective, the general RIS framework can be interpreted as that the informative tokens (\textit{e.g}., language tokens in previous methods) regarding the target object are leveraged as a key-value set to assign attention to target regions for vision query features. In other words, the role of the key-value set is a target information provider that guides the network on which regions to focus its attention. %, and the network predicts target regions based on the key-value information. %%%%%%%%%%%%%%%%%%%%%%%%%%% Require Rebutal Feedback %%%%%%%%%%%%%%%%%%%%%%%%%%%%%%%%%%
However, there is an inherent limitation of RIS task \textit{i.e}., target information is only given by a free-form language input, which has been noted in failure cases of prior works \cite{xu2023bridging, yu2023zero}.
 %Despite using advanced language tokens, such as visual-aware language tokens and LLM-generated tokens, these prior frameworks only use language-based tokens as a key-value set.
 Previous works using visual-aware language tokens as a key-value set have achieved performance improvements by leveraging the visual information into the language tokens.
  % This demonstrates that exploiting visual information is beneficial in compensating for the target information of the language input.
 %using these advanced language tokens as a key-value set is still not enough to ensure fine-grained segmentation, as shown in Figure \ref{_intro} (c).
Inspired by these works, we explore the following key questions: %\textit{How can we leverage the visual information more effectively for the effective key-value set to provide semantic target information to vision query features in the segmentation network for fine-grained segmentation?}
\textit{What is the effective key-value set for providing semantic target information to vision query features in the segmentation network? How can we leverage the visual information more effectively for fine-grained segmentation?} % How can we fine the effective key-value set that leverages the visual information

We propose a novel Visual Informative Part Attention (VIPA) framework to more effectively exploit visual contexts for referring image segmentation. VIPA framework aligns the query and key-value representations within the visual modality, thereby reducing the high-variance cross-modal projection and enhancing semantic consistency in the attention mechanism. Specifically, VIPA leverages the retrieved informative parts of visual contexts as a key-value set for the vision query features in the Transformer-based segmentation decoder; such informative parts of visual contexts are called \textit{Visual Expression} in this paper. We explore the linguistic context-relevant retrieved parts of visual tokens for visual expression, since the given linguistic context is inherently involved with target information. These informative parts of visual contexts contain more structural and semantic target information that is crucial for fine-grained segmentation. Hence, the visual expression robustly leads the network’s attention to the region of interest by providing semantic visual target contexts to the network on this task. 
 As shown in Fig. \ref{_intro}(c), the network’s attention by language-based tokens tends to miss the target regions or attend to even wide non-target regions. In contrast, our VIPA facilitates alignment between the network’s attention and the fine-grained region of interest.
 These observations indicate that our visual expression approach leverages visual information more effectively rather than the prior approach projecting visual information into language tokens. %using visual-aware language tokens. 
To the best of our knowledge, VIPA approach is the first to explicitly explore the informative parts of visual contexts as a key-value set for an advanced information provider in the attention mechanism of the referring image segmentation.

We also design a visual expression generator (VEG) module for informative visual expression. Firstly, this module leverages local-global linguistic contexts as cues to retrieve the informative visual tokens among all visual tokens, because most visual pixels in the image are not in the region of interest. The \verb|[CLS]| token and word tokens of language features contain the global (\textit{i.e}., sentence-level) context and the local (\textit{i.e.}, word-level) context of each word, respectively. By exploiting these local-global contexts, the visual expression can consider both the comprehensive context and the attribute contexts for the target object. Secondly, our module implements the visual context refinement step to ensure the informativeness of visual expression tokens. In this step, the retrieved visual tokens are refined to mitigate the possibility of being distracted by noise information and share visual attributes with each other for more attention to be devoted to semantic information. %to acquire visual contexts of the informative regions for devote more attention to semantic information. %this module refines informative visual tokens from the retrieved tokens in order to mitigate the possibility of being distracted by noise information among the retrieved tokens and to devote more attention to semantic information. Then, the refined informative tokens share their visual attributes with each other to acquire visual contexts of the informative regions. 
By doing so, the visual expression generated from the retrieved informative visual tokens serves as an effective key-value set in the Transformer-based segmentation decoder.  

The effectiveness of VIPA is validated through comprehensive visual analyses and extensive experiments across four public RIS benchmarks. %In addition, our method outperforms the existing Transformer-based RIS methods on all of four datasets. %thanks to the introduction of the visual expression as a key-value set in the Transformer-based referring image segmentation network. % Notably, without the LLM’s ability, our VIPA surpasses several LLM-based RIS models thanks to the introduction of the visual expression as a key-value set in the Transformer-based referring image segmentation network. 
In particular, Fig. \ref{ablation} (c) demonstrates that the visual expression provides key-value representations that are aligned in the visual feature space of the query, resulting in lower modality projection entropy compared to the language-based key-value set.

Our contributions can be summarized as follows:
\begin{itemize}
\item We propose a novel Visual Informative Part Attention (VIPA) framework for referring image segmentation, which leverages the informative parts of visual contexts as a key-value set for the vision query features in the Transformer-based segmentation decoder.
%to effectively provide semantic and structural target information in the Transformer-based segmentation decoder. 
Our approach is the first to explore the potential of visual expression in the attention mechanism of the referring segmentation.
\item We design a visual expression generator module, which retrieves informative visual tokens via local-global linguistic cues and refines them for mitigating the distraction by noise information and sharing the visual attributes. This module enables visual expression to consider comprehensive contexts and capture semantic visual contexts for fine-grained segmentation.
%\item We introduce a visual expression generation module that captures comprehensive contexts for the target region and refining  noise information.
\item VIPA consistently shows strong performance on four RIS benchmarks. The visual analysis of attention results clearly demonstrates the effectiveness of our approach. 
\end{itemize}

\section{Related works}
\label{sec:Related works}

\subsection{Referring Image Segmentation}
%여기서는 contrastive loss 기반의 방법, fusion 기반 방법, llm 기반 방법으로 설명하고 우리는 decoder 단에서 fusion 구조의 중요성에 focus해서 그 단을 건드리는 연구 할거다라는 플로우로 작성

Unlike single-modal semantic segmentation methods \cite{ding2020semantic, shuai2018toward, shim2023feedformer, Kang_2024_WACV, yu2024embedding, ravi2024sam} that operate on predefined categories, referring image segmentation (RIS) tackles the challenge of grounding unconstrained natural-language expressions to corresponding visual regions. Since the goal of RIS is to detect and segment the target object specified by language, a variety of approaches have focused on strengthening cross-modal alignment.
Contrastive learning has proven particularly effective for aligning visual and linguistic representations by pulling positive pairs closer while pushing negative pairs apart. CRIS \cite{wang2022cris} extends the contrastive loss used in CLIP to a pixel-level formulation, introducing a text-to-pixel contrastive loss. CrossVLT \cite{cho2023cross} applies contrastive loss at multiple encoder stages, utilizing rich intermediate visual–linguistic cues to further enhance alignment. CGFormer \cite{tang2023contrastive} incorporates a contrastive objective between learnable tokens and linguistic tokens, enabling the model to identify referent-related tokens and generate their corresponding segmentation masks.

In addition to contrastive-based alignment, several structural designs have been proposed to improve cross-modal feature interaction. Some studies \cite{li2018referring, hua2023multiple} fused language features and CNN-based visual features through concatenation. CMSA \cite{ye2019cross} concatenates vision–language features and adopts self-attention to capture long-range dependencies. 
Transformer-based methods\cite{ding2022vlt,zhu2022seqtr,kim2022restr,liu2022instance} perform fusion by applying cross-attention modules to the final features of the vision–language encoder, enabling cross-modal interaction between visual and linguistic representations.
However, these models typically perform fusion only on the final encoder features, failing to fully exploit rich intermediate information within the encoder. To address this limitation, early-fusion approaches \cite{cho2023cross,yang2022lavt} integrate cross-attention deeper inside the encoder, thereby improving cross-modal feature fusion. In addition, methods that follow the mask-decoder framework \cite{dingvlt,li2025bring, li2024bidirectional} generate query tokens from language-guided tokens and use them to extract mask features for final prediction.

More recent studies \cite{lai2024lisa, ren2024pixellm, rasheed2024glamm, xia2024gsva} employed large language models (LLMs) to improve language understanding. LISA \cite{lai2024lisa} was the first model to utilize the generated linguistic token by the LLM as a key-value in the segmentation decoder. Following its success, subsequent works \cite{wang2025segllm, yang2023lisa++} incorporate multiple LLM-generated tokens to further enrich language-guided reasoning. In parallel, several SAM-based frameworks \cite{liu2025refersam, xia2024gsva} have also been proposed to enhance cross-modal interaction in referring image segmentation.
However, although these methods generally demonstrate strong performance, they suffer from significant overhead due to the large model size and substantial computational cost.

Our work is motivated by the observation that effective cross-modal feature fusion is crucial for RIS performance. Rather than exploring where to place cross-attention, as in prior work, we instead examine the internal design of the cross-attention mechanism and its impact on cross-modal alignment and segmentation.

\subsection{Attention Mechanism in Referring Image Segmentation}
With the advancement of Transformers, cross-modal alignment methods based on the cross-attention mechanism have emerged in various vision-language tasks. In referring image segmentation, VLT \cite{ding2022vlt} first introduced a Transformer-based architecture to extract vision-aware language query vectors that capture diverse comprehension of language expression. They used these language vectors in the mask decoder for target segmentation.
% transformer의 발전으로 cross-attention을 기반으로 한 cross-modal alignment 방법들이 소개되었다. VLT[]는 vision guided attention을 이용해 vision 정보를 가진 language query vector들을 생성해내고 이를 transformer decoder에 input으로 하여 response를 생성하고, 이를 mask decoder에 활용하는 방법을 취하고 있다. Recent studies have explored Transformer-based RIS architectures where the vision features, as queries, reference the target-informative tokens as a key-value set in the segmentation decoder. 

Recent RIS studies have explored Transformer-based referring image segmentation (RIS) architectures in which visual features, as queries, reference target-informative tokens as a key–value set in the segmentation decoder. LAVT \cite{yang2022lavt} employed cross-attention within the vision encoder, which uses language features as a key–value set. CRIS \cite{wang2022cris} used the vanilla linguistic features as key-value elements in the Transformer decoder for segmentation. LQMFormer \cite{shah2024lqmformer} utilized learnable tokens, which are fine-tuned based on language expression, to extract diverse linguistic representations.
Several methods \cite{cho2023cross,hu2023beyond, tang2023contrastive, xu2023bridging, wangbarleria,liu2023multi, xu2024cmirnet, shang2022cross, xiao2025temporal} exploited the advanced linguistic features as a key-value set, which are extracted by bidirectional cross-attention to enhance the alignment between cross-modal information. CGFormer \cite{tang2023contrastive} used the advanced language tokens as keys-values by projecting visual information into learnable tokens conditioned on language features.
%LAVT[]는 vision encoder 내부에서 cross-attention구조를 활용하여, language feature를 key-value로 활용하여 fusion을 수행한다. 그리고 이를 encoder에서 element-wise addition하고, decoder에서 concatenate를 한번 더 해줌으로써 fusion을 수행한다. 그리고 이러한 계열로 CRIS \cite{wang2022cris} used the vanilla linguistic features as key-value elements하여 language information을 into vision feature하여 fusion 한다. ReSTR[]은 vision-langauge tokens을 concatenate하여 transformer encoder에 입력함으로써 fusion을 수행한다.
%ReSTR \cite{kim2022restr} exploited the concatenated vision–language tokens and fed them into a Transformer encoder to extract fused tokens. 
%Several methods \cite{cho2023cross,hu2023beyond, tang2023contrastive, xu2023bridging, wangbarleria,liu2023multi} exploited the advanced linguistic features as a key-value set, which are extracted by bidirectional cross-attention to enhance the alignment between cross-modal information. 
%LQMFormer \cite{shah2024lqmformer} utilized learnable tokens, which are fine-tuned based on language expression, to extract diverse linguistic representations.
%CGFormer[]는 learnable query tokens 들을 이용하여 visual information을 into tokens conditioning on language에 fusion한다.

Different from previous works, our framework leverages the retrieved informative visual tokens (\textit{i.e}., visual expression) as a key-value set for visual informative part attention in the Transformer-based segmentation decoder, rather than projecting visual information into language tokens. By directly providing semantically and structurally rich visual contexts of the fine-grained target regions, the visual expression robustly guides the network’s attention toward the region of interest.

\section{Method}

\begin{figure*}[t]
\includegraphics[width=\linewidth]{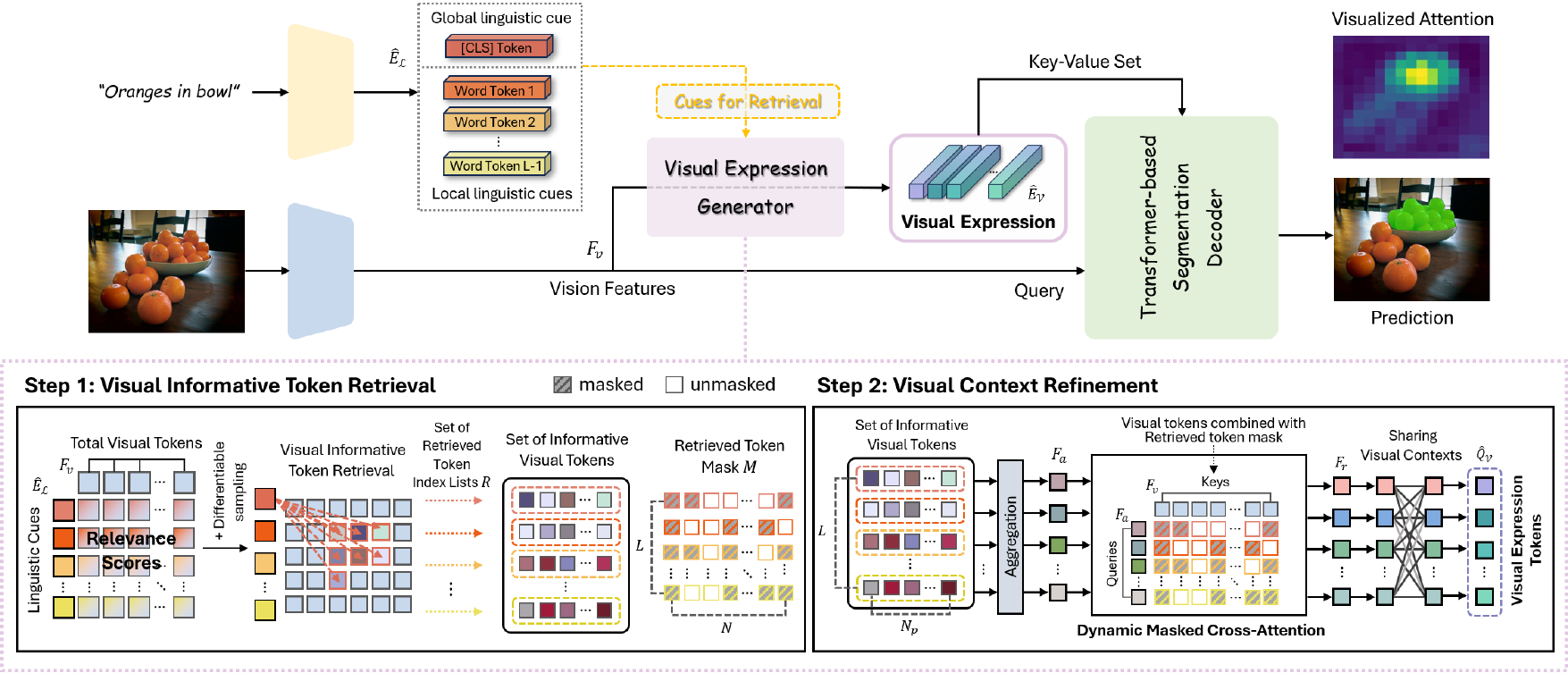}
\caption{Overview of Visual Informative Part Attention (VIPA) framework. VIPA robustly guides the network's attention to the region of interest by exploiting the visual expression generated from the retrieved informative parts of visual contexts.}
\label{_arch}
\end{figure*}

We propose a novel Visual Informative Part Attention (VIPA) framework for referring image segmentation, as illustrated in Fig. \ref{_arch}. We first describe a vision and language feature extraction, and then introduce a visual expression generator. Finally, we explain a segmentation decoder.

\subsection{Vision and Language Feature Extraction}
\label{encoder}
Given an image $\mathcal{I}$ and a linguistic expression $\mathcal{L}$ that consists of $L-1$ words, a vision encoder extracts the vision features ${{F}}_{i} \in \mathbb{R}^{H_{i}W_{i}\times C_{i}}$ at each stage $i\in\{1,2,3,4\}$ and a language encoder extracts the linguistic expression tokens ${{E}}_\mathcal{L}=[{l}_{cls}, {l}_1, ..., {l}_{L-1}] \in \mathbb{R}^{L \times D}$. Note that $H_{i}$, $W_{i}$, $C_{i}$ and $D$ denote the height, width, channel dimension of the feature maps at the $i^{th}$ vision stage, and the channel dimension of linguistic features. The first token ${l}_{cls}$ of linguistic expression features indicates a special \verb|[CLS]| token, which is the global representation that understands the linguistic expression at the sentence level. The word token ${l}_j$ indicates the local representation of $j^{th}$ word. The advanced linguistic expression tokens $\widehat{{E}}_\mathcal{L}$ are extracted by the cross-attention layers using the vision features as key-value to improve the comprehension for the language contexts.

\subsection{Visual Expression Generator}
\label{propose} 
To effectively guide the network to the region of interest, we produce the visual expression, which contains structural and semantic visual contexts. As shown in Fig. \ref{_arch}, the visual expression generator consists of the following two steps. 

\noindent
\textbf{Visual Informative Token Retrieval.} To retrieve the informative visual tokens, this step exploits local-global linguistic contexts as cues to consider both comprehensive context and attribute contexts. First, the vision features ${F}_{v}(={F}_4) \in \mathbb{R}^{N\times C}$ and the advanced local-global linguistic tokens $\widehat{{E}}_\mathcal{L}$ are embedded into the joint embedding space by the linear projection $\phi$, where $N$ is the total number of vision pixels. Then, the relevance score map ${S_c} \in \mathbb{R}^{L \times N}$ between the vision tokens and the linguistic tokens is computed as:
\begin{equation}\label{eq1}
    {{X}}= \phi^{\mathcal{V}}({F_v})\ ,\ {{Y}} = \phi^{\mathcal{L}}(\widehat{E}_\mathcal{L}) \ , \ {{S_c}} = \mathcal{C}({X}, {Y}),
\end{equation}
where $\mathcal{C}$ denote the cosine similarity function. To retrieve the informative visual tokens based on the higher relevance scores per linguistic cue, we employ the differentiable sampling technique \cite{jang2016categorical} during training, because the naive retrieval operation using the index list of the retrieved visual tokens is non-differentiable. 
\begin{equation}\label{eq2}
  S' = \mathtt{Softmax}(({S_c} + g) / \tau)\ ,  
  \ R = \mathcal{R}(S', r)\ ,
\end{equation}
\begin{equation}\label{eq3}
M = \mathtt{sampling}(R) \in \{0, 1\}^{L\times N} \ ,
\end{equation}
where $g$ and $\mathcal{R}$ denote a gumbel distribution and a relevance-based retrieval operation that retrieves the informative visual tokens based on the higher relevance scores per linguistic cue. $\tau$ is a learnable parameter. $r$ is a ratio of the retrieved tokens to total visual tokens. ${R}\in \mathbb{R}^{L \times N_p}$ is the set of the retrieved informative visual token index lists per linguistic token, where $N_p$ denotes the number of the retrieved visual tokens. The retrieved token mask $M \in \mathbb{R}^{L\times N}$, where `1' indicates a informative token and `0' is not in the region of interest, is obtained based on the set $R$ of the retrieved informative visual token index lists. The vision features ${F}_{v}$ and the retrieved token mask $M$ are passed to the next step.

To prevent the high relevance scores between the linguistic cues and the incorrect regions, the relevance score map $\textbf{s}\in\mathbb{R}^{1\times N}$ of the global linguistic token is supervised by a pixel contrastive loss: 
\begin{equation}\label{eq9}
    {l}oss_{c}=\begin{cases}-\mathrm{log} (\sigma(\textbf{s}_{z} )) & if\ z\in \mathcal{Z}^{+} \\  -\mathrm{log} (1-\sigma(\textbf{s}_{z} )) & if\ z\in \mathcal{Z}^{-}\end{cases} , 
\end{equation} 
where $\mathcal{Z}^{+}$ and $ \mathcal{Z}^{-}$ denote the set of the relevant pixels and irrelevant pixels for the ground truth target regions. $\sigma$ is a sigmoid function. The pixel contrastive loss {\cite{wang2022cris}} encourages that the relevant pixels are embedded closer together for high relevance score and the irrelevant pixels are embedded far apart for low relevance score.

\noindent
\textbf{Visual Context Refinement.} Rather than simply aggregating the retrieved information, we refine the informative visual tokens from the retrieved tokens to reduce the possibility of being distracted by noise information and to devote more attention to semantic information. 
In this step, the aggregated visual tokens ${F}_{a}\in \mathbb{R}^{L\times D}$ corresponding to each linguistic cue are first obtained by the summation of the retrieved visual tokens and the linear projection. Then, the visual tokens ${F}_{r}\in \mathbb{R}^{L\times D}$ are extracted by adaptively refining each aggregated visual token ${F}_{a}$ via the dynamic masked cross-attention mechanism for mitigating noise information and devoting more attention to semantic information as:
\begin{equation}\label{eq5}
      {F}_{a} = \mathtt{linear}(\sum^{N_p}(\mathtt{repeat}(F_v, L) \odot M) \in \mathbb{R}^{L\times D}, 
\end{equation}
\begin{equation}\label{eq7}
     \widehat{{F}} = \mathtt{MHCA}({F}_{a}, {F}_{v}, {M}) + {F}_{a},\
    {F}_r = \mathtt{MLP}(\widehat{{F}})+\widehat{{F}},
\end{equation}
where $\odot$ is element-wise multiplication, and $\mathtt{repeat}(f, x)$ indicates repeating the $f$ feature $x$ times to expand the shape.  $\mathtt{MHCA}(q, kv, m)$ is the multi-head cross-attention using $q$ as queries, $kv$ as key-value pairs and $m$ as masks. By using the retrieved token mask $M$ corresponding to each linguistic cue in the masked cross-attention, the intermediate visual tokens $\widehat{{F}} \in \mathbb{R}^{L\times D}$ per linguistic cue can capture semantic information from the informative visual tokens retrieved by the corresponding linguistic cue.

The visual expression tokens $\widehat{{E}}_\mathcal{V}=[{v}_{cls}, {v}_1, ..., {v}_{L-1}] \in \mathbb{R}^{L \times D}$ are finally produced by mutually sharing informative visual attributes to acquire visual contextual information as:
\begin{equation}\label{eq8}
    \widehat{{E}} = \mathtt{MHSA}({F}_{r}) + {F}_{r}\ ,\
    \widehat{{E}}_\mathcal{V} = \mathtt{MLP}(\widehat{{E}})+\widehat{{E}}\ ,
\end{equation}
where $\mathtt{MHSA}$ and $\widehat{E}$ indicate the multi-head self-attention, and the intermediate features, respectively. In this way, the visual expression captures semantic visual contexts for the fine-grained target segmentation.
%The normalized score map ${S}_{norm}$ is obtained by normalizing the whole relevance score map ${S_c}$ combined with the inverted mask ${\widehat{M}}$. The informative visual information per linguistic cue is aggregated by the normalized weighted sum to obtain $F_a$.
%Then, the refined visual tokens ${F}_{r}\in \mathbb{R}^{L\times D}$ are extracted by refining each aggregated visual token ${F}_{a}$ via the dynamic masked cross-attention mechanism to adaptively highlight the semantic information from the informative visual tokens, as follows: 

%\noindent
%\textbf{Visual relationship modeling.} The visual expression tokens $\widehat{{Q}}_\mathcal{V}=[\textbf{v}_{cls}, \textbf{v}_1, ..., \textbf{v}_{L-1}] \in \mathbb{R}^{L \times D}$ are produced by considering the visual relationship to mutually share each visual token's attributes and acquire the visual contextual information formulated as: % improving the visual understanding of the fine-grained target regions, formulated as: 
%\begin{equation}\label{eq8}
%    \widehat{{Q}} = \mathtt{MHSA}({F}_{r}) + {F}_{r}\ ,\
%    \widehat{{Q}}_\mathcal{V} = \mathtt{MLP}(\widehat{{Q}})+\widehat{{Q}}\ ,
%\end{equation}
%where $\mathtt{MHSA}$ and $\widehat{Q}$ indicate the multi-head self-attention, and the intermediate features, respectively. In this way, the visual expression capture the semantic visual contexts of the fine-grained target regions. %is endowed with the target-oriented visual guidance ability, which complements the linguistic guidance.

\begin{table*}[t]
\centering
\renewcommand{\arraystretch}{0.95}
\caption{Performance comparison with the state-of-the-art methods on four public referring image segmentation datasets. LLM-based models are marked in \textcolor{Grey}{gray}. $^\dagger$ indicates models trained on multiple RefCOCO series datasets with removed validation and testing images to prevent data leakage. For ReferIt dataset, only ReferIt training set is used. } 
\resizebox{\textwidth}{!}{\small
\begin{tabular}{lcccccccccccc}
\toprule[1.4pt]
  \multirow{2}{*}{\textbf{Method}} &\multirow{2}{*}{\textbf{Publication}} &\multirow{1}{*}{\textbf{Vision}}&\multirow{1}{*}{\textbf{Language}}& \multicolumn{3}{c}{\textbf{RefCOCO}} & \multicolumn{3}{c}{\textbf{RefCOCO+}} & \multicolumn{2}{c}{\textbf{RefCOCOg}}&\textbf{ReferIt} \\ [-0.1em] \cmidrule(lr){5-7} \cmidrule(lr){8-10} \cmidrule(lr){11-12} \cmidrule(lr){13-13}\\[-1.1em]
&&\textbf{Model}& \textbf{Model} &\multicolumn{1}{c}{\textit{val}} & \multicolumn{1}{c}{\textit{test A}} &  \textit{test B} & \multicolumn{1}{c}{\textit{val}} & \multicolumn{1}{c}{\textit{test A}} &  \textit{test B} & \multicolumn{1}{c}{\textit{val}} & \textit{test} & \textit{test} \\ \midrule

\multicolumn{12}{l}{\textit{{LLM-based RIS methods + additional vision-language training datasets}}}\\ \midrule
\multicolumn{1}{l}{\textcolor{Grey}{LISA-7B} \textcolor{Grey}{\cite{lai2024lisa}}} & \textcolor{Grey}{CVPR 2024} & \textcolor{Grey}{SAM-H}  & \textcolor{Grey}{Vicuna-7B}&\textcolor{Grey}{74.9} & \textcolor{Grey}{79.1} & \textcolor{Grey}{72.3} & \textcolor{Grey}{65.1} & \textcolor{Grey}{70.8} & \textcolor{Grey}{58.1} & \textcolor{Grey}{67.9} & \textcolor{Grey}{70.6}&\textcolor{Grey}{-} \\
 \multicolumn{1}{l}{\textcolor{Grey}{PixelLM} \textcolor{Grey}{\cite{ren2024pixellm}}} &\textcolor{Grey}{CVPR 2024}  & \textcolor{Grey}{CLIP-VIT-L} &\textcolor{Grey}{Vicuna-7B} &\textcolor{Grey}{73.0} & \textcolor{Grey}{76.5} & \textcolor{Grey}{68.2} & \textcolor{Grey}{66.3} &\textcolor{Grey}{71.7} & \textcolor{Grey}{58.3} & \textcolor{Grey}{69.3} & \textcolor{Grey}{70.5} &\textcolor{Grey}{-} \\
\multicolumn{1}{l}{\textcolor{Grey}{GSVA-7B} \textcolor{Grey}{\cite{xia2024gsva}}}&  \textcolor{Grey}{CVPR 2024} &\textcolor{Grey}{SAM-H}&\textcolor{Grey}{Vicuna-7B}& \textcolor{Grey}{77.2} &\textcolor{Grey}{78.9} &\textcolor{Grey}{73.5}& \textcolor{Grey}{65.9}& \textcolor{Grey}{69.6}& \textcolor{Grey}{59.8}& \textcolor{Grey}{72.7}& \textcolor{Grey}{73.3}&\textcolor{Grey}{-}\\
\multicolumn{1}{l}{\textcolor{Grey}{GLaMM} \textcolor{Grey}{\cite{rasheed2024glamm}}}  &\textcolor{Grey}{CVPR 2024}&\textcolor{Grey}{SAM-H} & \textcolor{Grey}{Vicuna-7B}&\textcolor{Grey}{79.5} &\textcolor{Grey}{83.2}& \textcolor{Grey}{76.9}& \textcolor{Grey}{72.6}& \textcolor{Grey}{78.7}& \textcolor{Grey}{64.6}& \textcolor{Grey}{74.2}& \textcolor{Grey}{74.9}&\textcolor{Grey}{-}\\
\multicolumn{1}{l}{\textcolor{Grey}{SAM4MLLM-7B} \textcolor{Grey}{\cite{chen2025sam4mllm}}}&\textcolor{Grey}{ECCV 2024}&\textcolor{Grey}{SAM-XL}&\textcolor{Grey}{Qwen-VL-7B}& \textcolor{Grey}{76.2}& \textcolor{Grey}{80.1}& \textcolor{Grey}{72.0} &\textcolor{Grey}{71.2}&\textcolor{Grey}{75.9} &\textcolor{Grey}{64.3}& \textcolor{Grey}{74.2}& \textcolor{Grey}{74.3} &\textcolor{Grey}{-}\\
\multicolumn{1}{l}{\textcolor{Grey}{Text4Seg-7B} \textcolor{Grey}{\cite{lan2024text4seg}}}&\textcolor{Grey}{ICLR 2025}& \textcolor{Grey}{SAM-H}& \textcolor{Grey}{Vicuna-7B} & \textcolor{Grey}{79.3}& \textcolor{Grey}{81.9}& \textcolor{Grey}{76.2}& \textcolor{Grey}{72.1}& \textcolor{Grey}{77.6}& \textcolor{Grey}{66.1}& \textcolor{Grey}{72.1}& \textcolor{Grey}{73.9}& \textcolor{Grey}{-}\\
\multicolumn{1}{l}{\textcolor{Grey}{SegLLM} \textcolor{Grey}{\cite{wangsegllm}}}&\textcolor{Grey}{ICLR 2025} &\textcolor{Grey}{SAM-H}& \textcolor{Grey}{Vicuna-7B}&\textcolor{Grey}{80.2}& \textcolor{Grey}{81.5}& \textcolor{Grey}{75.4}& \textcolor{Grey}{70.3}& \textcolor{Grey}{73.0}& \textcolor{Grey}{62.5}& \textcolor{Grey}{72.6}& \textcolor{Grey}{73.6}&\textcolor{Grey}{-}\\
\multicolumn{1}{l}{\textcolor{Grey}{F-LMM} \textcolor{Grey}{\cite{wu2024f}}}&\textcolor{Grey}{CVPR 2025}&\textcolor{Grey}{SAM-L}&\textcolor{Grey}{LLaMA2-7B}&\textcolor{Grey}{75.2} &\textcolor{Grey}{-}&\textcolor{Grey}{-}&\textcolor{Grey}{63.7}&\textcolor{Grey}{-}&\textcolor{Grey}{-}& \textcolor{Grey}{67.1}&\textcolor{Grey}{-}&\textcolor{Grey}{-}\\
\midrule%\cline{2-13}
%\\[-1.2em]

 %& \multicolumn{1}{|l}{LAVT \cite{yang2022lavt}} &Swin-B &BERT-B & \multicolumn{1}{c}{{72.73}} & \multicolumn{1}{c}{{75.82}} &  {68.79} & \multicolumn{1}{c}{62.14} & \multicolumn{1}{c}{{68.38}} &  {55.10} & \multicolumn{1}{c}{{61.24}} &  {62.09}&- \\ 
 %& CoupAlign \cite{zhang2022coupalign} & NeurIPS '22 &Swin-B &BERT-B& 74.70 & {77.76} & 70.58 & 62.92 & 68.34 & 56.69 & 62.84 & 62.22 & -\\
 \multicolumn{12}{l}{\textit{Transformer-based RIS methods (oIoU)}}\\ \midrule
  \multicolumn{1}{l}{VLT \cite{ding2022vlt}} &TPAMI 2023 & Swin-B&BERT-B&\multicolumn{1}{c}{72.96} & \multicolumn{1}{c}{75.96} & 69.60 & 63.53 & 68.43 & 56.92 & 63.49 & {66.22} & -\\
% & RefSegformer \cite{wu2024towards} & & & \multicolumn{1}{c}{73.22} & \multicolumn{1}{c}{75.64} & 70.09 & \multicolumn{1}{c}{63.50} & \multicolumn{1}{c}{68.69} & 55.44 & \multicolumn{1}{c}{62.56} & 63.07 & 58.48\\
   \multicolumn{1}{l}{ReLA \cite{liu2023gres}} & CVPR 2023 &Swin-B&BERT-B& 73.82 & 76.48 & 70.18 & 66.04 & 71.02 & 57.65 & 65.00 & 65.97 & -\\
 %& \multicolumn{1}{|l}{SADLR  \cite{yang2023semantics}} & Swin-B&BERT-B&\multicolumn{1}{c}{74.24} & \multicolumn{1}{c}{76.25} & 70.06 & \multicolumn{1}{c}{64.28} & \multicolumn{1}{c}{69.09} & 55.19 & \multicolumn{1}{c}{63.60} & 63.56 & 61.16 \\
  \multicolumn{1}{l}{DMMI \cite{hu2023beyond}} & ICCV 2023  &Swin-B&BERT-B& \multicolumn{1}{c}{74.13} & \multicolumn{1}{c}{77.13} & 70.16 & \multicolumn{1}{c}{63.98} & \multicolumn{1}{c}{69.73} & 57.03 & \multicolumn{1}{c}{63.46} & 64.19 & -\\
\multicolumn{1}{l}{LQMFormer \cite{shah2024lqmformer}}&CVPR 2024&Swin-B&BERT-B&74.16 &76.82& 71.04& 65.91& {71.84}& 57.59& 64.73& 66.04& -\\
  \multicolumn{1}{l}{CGFormer \cite{tang2023contrastive}} &CVPR 2023 & Swin-B&BERT-B&\multicolumn{1}{c}{74.75} & \multicolumn{1}{c}{77.30} & 70.64 & \multicolumn{1}{c}{64.54} & \multicolumn{1}{c}{71.00} & 57.14 & \multicolumn{1}{c}{64.68} & 65.09 & 73.36\\
  \multicolumn{1}{l}{MagNet \cite{chng2024mask}}&CVPR 2024&Swin-B&BERT-B&75.24& 78.24& 71.05 &66.16& 71.32& 58.14& 65.36& 66.03&-\\
   \multicolumn{1}{l}{\cellcolor{lavender(web)}{\textbf{\textsc{\method} (Ours)}}} & \cellcolor{lavender(web)}{-}&\cellcolor{lavender(web)}{Swin-B} &\cellcolor{lavender(web)}{BERT-B} &\cellcolor{lavender(web)}{\textbf{75.84}} & \cellcolor{lavender(web)}{\textbf{78.58}}&\cellcolor{lavender(web)}{\textbf{72.20}}&\cellcolor{lavender(web)}{\textbf{66.91}}&\cellcolor{lavender(web)}{\textbf{72.44}}&\cellcolor{lavender(web)}{\textbf{60.15}}&\cellcolor{lavender(web)}{\textbf{65.93}}&\cellcolor{lavender(web)}{\textbf{67.37}}&\cellcolor{lavender(web)}{\textbf{74.55}}\\
  \midrule
  %X-Decoder \cite{zou2023generalized}&DaViT-L&Transformer&-& -& -& -& - &- &64.6 &- &-\\
 % SEEM \cite{zou2024segment} & DaViT-L &&-& - &-& - &- &- &65.6 &-& -\\
 \multicolumn{1}{l}{PolyFormer-B$^\dagger$ \cite{liu2023polyformer}}&CVPR 2023&Swin-B&BERT-B&74.82 &76.64& 71.06& 67.64& 72.89& 59.33 &67.76& 69.05 &71.91\\
 \multicolumn{1}{l}{MagNet$^\dagger$ \cite{chng2024mask}}&CVPR 2024&Swin-B&BERT-B&76.55 &78.27& 72.15 &68.10& 73.64 &61.81 &67.79& 69.29&-\\%{75.24} &{78.24}&{71.05}& {66.16}& 71.32& {58.14} & {65.36}& 66.03 &-\\
% & \multicolumn{1}{|l}{\cellcolor{lavender(web)}{\textbf{\textsc{\method} (Ours)}}} & \cellcolor{lavender(web)}{Swin-B}&\cellcolor{lavender(web)}{BERT-B}&\cellcolor{lavender(web)}{\textbf{75.35}} & \cellcolor{lavender(web)}{{77.97}} & \cellcolor{lavender(web)}{\textbf{71.94}} & \cellcolor{lavender(web)}{\textbf{66.70}} & \cellcolor{lavender(web)}{\textbf{72.08}} & \cellcolor{lavender(web)}{\textbf{59.85}} & \cellcolor{lavender(web)}{\textbf{65.78}} & \cellcolor{lavender(web)}{\textbf{66.93}} & \cellcolor{lavender(web)}{\textbf{63.49}} \\
 \multicolumn{1}{l}{\cellcolor{lavender(web)}{\textbf{\textsc{\method}$^\dagger$ (Ours)}}}& \cellcolor{lavender(web)}{-}&\cellcolor{lavender(web)}{Swin-B} &\cellcolor{lavender(web)}{BERT-B} &\cellcolor{lavender(web)}{\textbf{78.14}} & \cellcolor{lavender(web)}{\textbf{80.56}}&\cellcolor{lavender(web)}{\textbf{75.24}}&\cellcolor{lavender(web)}{\textbf{70.15}}&\cellcolor{lavender(web)}{\textbf{75.22}}&\cellcolor{lavender(web)}{\textbf{63.34}}&\cellcolor{lavender(web)}{\textbf{70.01}}&\cellcolor{lavender(web)}{\textbf{71.77}}&\cellcolor{lavender(web)}{\textbf{74.55}}\\
 \midrule
 \multicolumn{12}{l}{\textit{Transformer-based RIS methods (mIoU)}}\\ \midrule
CGFormer \cite{tang2023contrastive} & CVPR 2023&Swin-B&BERT-B&76.93& 78.70& 73.32& 68.56& 73.76& 61.72& 67.57& 67.83& 66.42\\
DETRIS-B \cite{huang2025densely} & AAAI 2025& DINOv2-B&CLIP&76.0 &78.2& 73.5& 68.9 &74.0& 61.5 &67.9 &68.1& -\\
\multicolumn{1}{l}{\cellcolor{lavender(web)}{\textbf{\textsc{\method} (Ours)}}}&\cellcolor{lavender(web)}{-}&\cellcolor{lavender(web)}{Swin-B}&\cellcolor{lavender(web)}{BERT-B} &\cellcolor{lavender(web)}{\textbf{78.68}}& \cellcolor{lavender(web)}{\textbf{79.93}}&\cellcolor{lavender(web)}{\textbf{75.98}}&\cellcolor{lavender(web)}{\textbf{70.42}}&\cellcolor{lavender(web)}{\textbf{74.64}}&\cellcolor{lavender(web)}{\textbf{62.96}}&\cellcolor{lavender(web)}{\textbf{69.98}}&\cellcolor{lavender(web)}{\textbf{70.94}}&\cellcolor{lavender(web)}{\textbf{67.88}}\\
PolyFormer-B$^\dagger$ \cite{liu2023polyformer} &CVPR 2023&Swin-B&BERT-B& 75.96 &77.09 &73.22& 70.65& 74.51& 64.64& 69.36& 69.88&65.98\\
EEVG$^\dagger$ \cite{chen2024efficient} &ECCV 2024&ViT-B&BERT-B& 78.23& 79.27& 76.58& 69.04& 72.65& 62.33 &69.15& 70.01&-\\
\multicolumn{1}{l}{\cellcolor{lavender(web)}{\textbf{\textsc{\method} $^\dagger$ (Ours)}}}&\cellcolor{lavender(web)}{-}&\cellcolor{lavender(web)}{Swin-B}&\cellcolor{lavender(web)}{BERT-B} &\cellcolor{lavender(web)}{\textbf{80.97}}& \cellcolor{lavender(web)}{\textbf{82.63}}&\cellcolor{lavender(web)}{\textbf{78.53}}&\cellcolor{lavender(web)}{\textbf{73.69}}&\cellcolor{lavender(web)}{\textbf{77.58}}&\cellcolor{lavender(web)}{\textbf{67.44}}&\cellcolor{lavender(web)}{\textbf{73.60}}&\cellcolor{lavender(web)}{\textbf{74.97}}&\cellcolor{lavender(web)}{\textbf{67.88}}\\
\bottomrule[1.4pt]
\end{tabular}} 
\label{tab:table1}
\end{table*}

\subsection{Segmentation Decoder with Visual Informative Part Attention}
\label{decoder} 
The visual expression $\widehat{{E}}_\mathcal{V}$ generated from the retrieved informative visual tokens is used as a key-value set, which serves as the target information provider in the Transformer-based segmentation decoder. In particular, the visual expression provides key-value representations that are aligned with the visual feature space of the query. This minimizes modality gap between the query and key-value spaces than using language-based key-value representations. Thus, VIPA enhances semantic consistency in the attention mechanism.

  At each stage of the segmentation decoder, the cross-attention layer, which uses vision features as a query and visual expression tokens as a key-value set, is employed to highlight the target regions, as follows: 
    \begin{equation}\label{eq10}
     {{F}_o} = \mathtt{MHCA}({F}_{v}, \widehat{{E}}_\mathcal{V}) + {F}_{v},\
    {F}_d = \mathtt{MLP}({{F}_o})+{{F}_o},
\end{equation}
where $\mathtt{MHCA}(q, kv)$ denotes the multi-head cross-attention using $q$ as a query, $kv$ as a key-value set.
  The decoder can focus its attention on the fine-grained regions of interest thanks to the guidance by the informative parts of visual contexts.
  The vision decoder features ${F}_d$ are then upsampled and concatenated with the corresponding vision encoder features to feed into the next decoder stage. The final segmentation map is projected to a binary class mask by a linear projection layer. The binary cross-entropy loss and the dice loss are used for the network training.

\section{Experiments}
\label{experiments}
\subsection{Implementation Details} 
\noindent
\textbf{Settings.} Our method was implemented in PyTorch \cite{paszke2019pytorch}. The vision encoder is Swin-B \cite{liu2021swin} initialized with the pretrained weight on ImageNet-22K \cite{krizhevsky2012imagenet}. The language encoder is BERT-base \cite{devlin2018bert} initialized with the official pretrained weight of the uncased version. The decoder was randomly initialized. We trained models for 40 epochs with 16 batch size on 24G RTX4090 GPUs. We used the AdamW \cite{loshchilov2017decoupled} optimizer with initial learning rate of 3e-5 and adopted the polynomial learning rate decay scheduler. The input image resolution was 480$\times$480. The maximum sequence length was set to 21 words including the \verb|[CLS]| token for all datasets. %More details for settings are in supplementary materials.

\begin{figure*}[t]
\centering
\includegraphics[width=\linewidth]{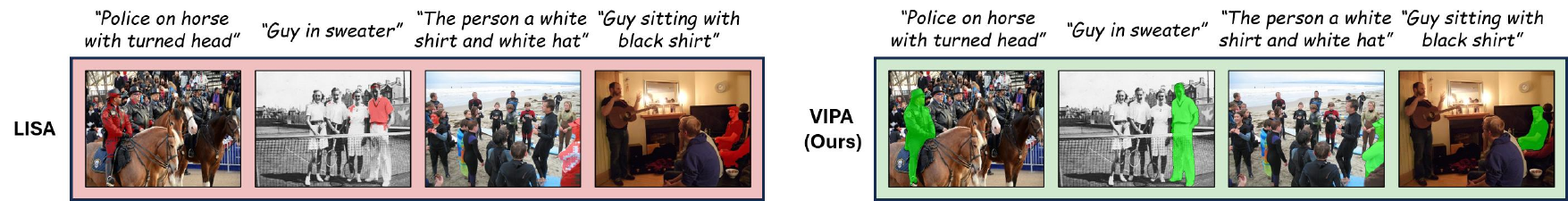} 
\caption{Qualitative comparison with the LLM-based RIS model \cite{lai2024lisa} on RefCOCO+.}
\label{lisa} 
\end{figure*}

\noindent
\textbf{Datasets.} RefCOCO \cite{yu2016modeling}, RefCOCO+ \cite{yu2016modeling}, RefCOCOg \cite{mao2016generation, nagaraja2016modeling} and ReferIt \cite{kazemzadeh2014referitgame} are widely used datasets for referring image segmentation. RefCOCO contains 19,994 images with 142,209 language expressions for 50,000 objects, and RefCOCO+ contains 19,992 images with 141,564 expressions for 49,856 objects. 
The expressions in RefCOCO+ do not include words about absolute locations, which makes it more challenging than RefCOCO. RefCOCOg is the most challenging dataset, which has more complex and longer expressions. 
It contains 26,711 images with 104,560 language expressions for 54,822 objects. In addition, ReferIt contains 19,997 images with 130,364 language expression for 99,296 objects.
%collected from the MSCOCO dataset, Each expression in RefCOCO and RefCOCO+ consists of 3.5 words on average. RefCOCO and RefCOCO+ contain 3.9 objects of the same category per image on average. RefCOCOg collected from & containing 8.4 words on average

\noindent
\textbf{Evaluation metrics.} Following prior works, we adopted the overall intersection-over-union (oIoU), mean intersection-over-union (mIoU), and precision at 0.5 and 0.7 thresholds. 
 The oIoU is the ratio between the total intersection regions and the total union regions of all test samples. The mIoU is the average of IoUs between the predicted mask and the ground truth of all test samples. The precision is the percentage of test samples that have a IoU score higher than a threshold.

\begin{table}[t]
\centering
\renewcommand{\arraystretch}{}
\caption{Comparison of more recent transformer-based methods using BEiT3 \cite{wang2023image} as encoder.}
\resizebox{\linewidth}{!}{\small
\begin{tabular}{lccccccccc}
\toprule[1.4pt]
\multirow{2}{*}{\textbf{Method}} & \multirow{2}{*}{\textbf{Publication}}&\multicolumn{3}{c}{\textbf{RefCOCO}} &\multicolumn{3}{c}{\textbf{RefCOCO+}} &\multicolumn{2}{c}{\textbf{RefCOCOg}} \\ [-0.2em]
\cmidrule(lr){3-5}\cmidrule(lr){6-8}\cmidrule(lr){9-10}
&& \textit{val}& \textit{test A} & \textit{test B} & \textit{val}& \textit{test A}&\textit{test B}&\textit{val} & \textit{test}\\
\midrule
C3VG \cite{dai2025multi}&AAAI'25&81.37& 82.93& 79.12& 77.05& 79.61& 72.40 &76.34& 77.10\\
DeRIS-B \cite{dai2025deris} &ICCV'25& 81.99& 82.97 &80.14& 75.62& 79.16& 71.63 &76.30& 77.15\\
\multicolumn{1}{l}{\cellcolor{lavender(web)}{\textbf{\textsc{\method} (Ours)}}} &\cellcolor{lavender(web)}{-}&\cellcolor{lavender(web)}{\textbf{82.41}}& \cellcolor{lavender(web)}{\textbf{83.64}}&\cellcolor{lavender(web)}{\textbf{80.57}}&\cellcolor{lavender(web)}{\textbf{77.36}}&\cellcolor{lavender(web)}{\textbf{80.19}}&\cellcolor{lavender(web)}{\textbf{72.48}}&\cellcolor{lavender(web)}{\textbf{76.72}}&\cellcolor{lavender(web)}{\textbf{77.45}}\\
\bottomrule[1.4pt]
\end{tabular}}  
\label{tab:table2}
\end{table}

\begin{table}[t]\centering 
\renewcommand{\arraystretch}{}
\caption{Computational comparison with recent methods on RefCOCOg \textit{val}. Inference times are measured in Brain Floating Point 16 (BF16) for LLM-based models and Floating Point 32 (FP32) for Transformer-based models. LLM-based models are marked in \textcolor{Grey}{gray}. $\ddagger$ indicates models trained with additional vision-language datasets. $\dagger$ indicates models trained on multiple RefCOCO series datasets.}
\resizebox{\linewidth}{!}{
\small
\begin{tabular}{L{2.5cm}C{1.4cm}C{1.4cm}C{2.8cm}}
    \toprule[1.3pt]
    {\textbf{Method}}& {\textbf{GFLOPs}}&  \textbf{oIoU (\%)}&\textbf{Inference Time (ms)}\\
    \midrule
    \textcolor{gray}{GLaMM-7B$^{\ddagger}$ \cite{rasheed2024glamm}}& \textcolor{gray}{-} &\textcolor{gray}{74.2}&\textcolor{gray}{786 (BF16)}\\
    \textcolor{gray}{LISA-7B$^{\ddagger}$ \cite{lai2024lisa}}& \textcolor{gray}{13,026} &\textcolor{gray}{67.9}&\textcolor{gray}{368 (BF16)}\\%&\textcolor{gray}{71.2}
    \cellcolor{lavender(web)}{{\textbf{VIPA$^{\dagger}$ (Ours)}}}&\cellcolor{lavender(web)}{431}&\cellcolor{lavender(web)}{\textbf{70.01}}&\cellcolor{lavender(web)}{82 (FP32)}\\
    DMMI \cite{hu2023beyond}&392&63.46&74 (FP32)\\ % &66.48
    CGFormer \cite{tang2023contrastive}&950&64.68&105 (FP32)\\ %&67.57
    \cellcolor{lavender(web)}{{\textbf{VIPA (Ours)}}}&\cellcolor{lavender(web)}{431}&\cellcolor{lavender(web)}{\textbf{65.93}}&\cellcolor{lavender(web)}{82 (FP32)}\\ %\cellcolor{lavender(web)}{\textbf{69.98}}
    \bottomrule[1.3pt]
    \end{tabular}}
\label{flops}
\end{table}

\begin{table}[t]
\centering
\renewcommand{\arraystretch}{}
\caption{Comparison for generalization setting using mIoU.}
\resizebox{\linewidth}{!}{\small
\begin{tabular}{L{2.6cm}C{1.5cm}C{1.55cm}C{1.5cm}C{1.5cm}}
\toprule[1.4pt]
\multirow{2}{*}{\textbf{Method}}& \multicolumn{2}{c}{\textbf{RefCOCOg \textit{val}}}&\multicolumn{2}{c}{\textbf{RefCOCOg \textit{test}}}\\
    \cmidrule(lr){2-3} \cmidrule(lr){4-5}  \\ [-1.3em]
&\textit{seen}&\textit{unseen}&\textit{seen}&\textit{unseen}\\ 
         \midrule
         CRIS \cite{wang2022cris} &58.64 &42.63& 59.68& 38.88\\ 
         LAVT \cite{yang2022lavt}  &60.16& 42.33& 60.37& 41.38\\
         CGFormer \cite{tang2023contrastive} &65.60 &46.11& 65.67& 42.31\\
         \cellcolor{lavender(web)}{{\textbf{\textsc{VIPA} (Ours)}}}&\cellcolor{lavender(web)}{\textbf{66.52}}&\cellcolor{lavender(web)}{\textbf{46.74}}&\cellcolor{lavender(web)}{\textbf{66.93}}&\cellcolor{lavender(web)}{\textbf{43.06}}\\
\bottomrule[1.4pt]
\end{tabular}}  
\label{tab:unseen}
\end{table}

\subsection{Comparison with Transformer-based RIS Methods} 
In Table \ref{tab:table1}, we evaluated our approach with Transformer-based RIS methods on four benchmarks. Our method consistently showed strong performance on all splits of all datasets. In a single dataset setting, our VIPA outperformed the state-of-the-art methods on all dataset splits. In a multiple dataset setting, our method surpassed MagNet with remarkable improvements of 2.23\% and 1.72\% on average for RefCOCO and RefCOCO+, respectively. Especially, our VIPA showed considerable margins of 2.22\% and 2.48\% on each split of RefCOCOg, most challenging benchmark.
For more comprehensive evaluation, we evaluated our method using mIoU. Compared to the state-of-the-art method CGFormer, our model significantly improved mIoU performance by 1.84\% on average on four benchmarks. 

In Table \ref{tab:table2}, we experimented with the powerful encoder \cite{wang2023image} for fair comparison with more recent methods. Our method showed competitive performance and can be successfully applied regardless of the specific encoder structure. Furthermore, in Table \ref{flops}, \textsc{\method} showed higher oIoU accuracy with comparable computations to DMMI and with 54.6\% less GFLOPs than CGFormer on the most challenging dataset. These results demonstrate the effectiveness of our visual informative part attention framework. 

 In addition, we evaluated the generalizability of our framework compared to existing methods. In referring image segmentation, the ability to understand the visual context within the image is particularly crucial for achieving strong generalizability. In Table \ref{tab:unseen}, we experimented with the generalization setting \cite{tang2023contrastive}, where only the language descriptions for the seen target object classes are given during training and the model is not trained with the ground truth masks for the unseen target object classes. Under this setting, VIPA surpassed the existing methods and consistently showed performance improvements on both seen and unseen sets. These results suggest that our method achieves better generalization ability than previous RIS methods by learning semantic visual contexts via the visual expression.

\subsection{Comparison with LLM-based RIS Methods} 
It is actually unfair to compare our model with LLM-based RIS models for the following reasons: (1) They adopt large-sized vision models and LLMs. (2) They use a strong segmentation model (SAM), which is trained with a large amount of segmentation datasets. (3) They are trained with large-scale vision-language grounding datasets.
Despite the unfair comparison, we conducted comparison with LLM-based RIS models for further analysis. In Table \ref{flops}, our model showed competitive performance compared to LLM-based models by effectively providing structural and semantic visual contexts via the visual expression. In addition, our model showed significantly lower computations and was 5$\sim$10 times faster than LLM-based models. This result supports the importance of continuing to study transformer-based RIS methods in order to consider in terms of both performance and efficiency. Furthermore, we compared segmentation results in Fig. \ref{lisa}. Our model showed precise segmentation, whereas LISA segmented only some part of a target object or segment even non-target regions. These results indicate that our model has a stronger ability to guide the network's attention to the fine-grained region of interest, compared to the LLM-based models.

\begin{table}[t]\centering
    \renewcommand{\arraystretch}{1.15}
    \captionsetup[table]{hypcap=false}
      \captionof{table}{Ablation for the effectiveness of the visual informative part attention framework. LE: Linguistic Expression tokens. VE: Visual Expression tokens (Ours). \colorbox{beige}{\rule[0.2em]{0pt}{0.1em} } 
       are ablation methods that leverage language-based tokens as key-value components in the segmentation decoder. Advanced LE indicates a method using visual-aware language tokens extracted by cross-attention with the vision features. \colorbox{lavender(web)}{\rule[0.2em]{0pt}{0.1em}  } is our method that leverages the visual expression, generated from the retrieved informative parts of visual contexts.%, as key-value components in the decoder.
      % is a model with target-informative visual guidance only. \colorbox{aliceblue}{\rule[0.2em]{0pt}{0.1em}  } is a model using all visual information as visual guidance. \colorbox{lavender(web)}{\rule[0.2em]{0pt}{0.1em}  } is our full model.
      }
        \resizebox{\linewidth}{!}{
        \small
        \begin{tabular}{ccccccccc}
        \toprule[1.4pt]
        \multirow{2}{*}{\textbf{Key-Value Set}}&\multicolumn{4}{c}{\textbf{{RefCOCO+  \textit{val}}}}&\multicolumn{4}{c}{\textbf{RefCOCOg  \textit{val}}}\\ [-0.1em]
         \cmidrule(lr){2-5} \cmidrule(lr){6-9}  \\ [-1.5em]
        %\\[-0.8em]
        % &{P@0.5}&{P@0.7}&{P@0.9}&{mIoU}&{oIoU}&{P@0.5}&{P@0.7}&P@0.9&mIoU&oIoU&{P@0.5}&{P@0.7}&{P@0.9}&{mIoU}&{oIoU}\\
        &{P@0.5}&{P@0.7}&mIoU&oIoU&{P@0.5}&{P@0.7}&{mIoU}&{oIoU}\\
        \midrule
        \cellcolor{beige}Vanilla LE  &
        73.77&64.89& 63.93&62.44  
        &72.84	&59.98&	62.80&	61.74\\%&35.01&28.40&22.95
        
        \cellcolor{beige}Advanced LE &
        75.26&66.49&66.14&64.16
        & 74.35&61.49& 65.06	&63.52\\ % &36.04&30.12&24.77
        
        %   \cellcolor{celadon}All pixels&86.17	&77.40&	36.73&	75.65& 74.36& 75.81&67.28&30.89&66.97&65.24&74.85  &62.77   &25.91  &66.02   &65.27\\
        \cellcolor{lavender(web)}VE  	&\cellcolor{lavender(web)}\textbf{78.64}&\cellcolor{lavender(web)}\textbf{69.62}&\cellcolor{lavender(web)}\textbf{70.42}&\cellcolor{lavender(web)}\textbf{66.91}&\cellcolor{lavender(web)}\textbf{77.03}&\cellcolor{lavender(web)}\textbf{65.26}& \cellcolor{lavender(web)}\textbf{69.98}  &\cellcolor{lavender(web)}\textbf{65.93}\\ 
        % &\cellcolor{lavender(web)}40.43&\cellcolor{lavender(web)}33.18&\cellcolor{lavender(web)}28.09

        % \cellcolor{lavender(web)}Advanced LE & \cellcolor{lavender(web)}VE  &\textbf{86.71}&\textbf{78.30}&\textbf{37.24}&\textbf{76.97}&\textbf{75.35}&\textbf{77.13}&\textbf{69.05}&\textbf{32.94}&\textbf{68.63}&\textbf{66.70}&\textbf{76.13}&\textbf{64.60}	&\textbf{27.87}&\textbf{67.85}&\textbf{66.93}\\
        \bottomrule[1.4pt]
      \end{tabular}}
      \label{tab:abltable1} 
\end{table}

\subsection{Ablation Studies}
In ablation setting, our default model and ablation models are trained on each single dataset, not multiple datasets. For fair comparisons, all ablation models are based on our network and we added more layers into  ablation models to maintain the model size similar to our default model.

\noindent
\textbf{Effectiveness of Visual Expression.}
In Table \ref{tab:abltable1}, we conducted experiments to validate the effectiveness of exploiting the visual expression as the component of the key-value. %Compared to `Vanilla LE' method that uses the vanilla language encoder features as key-value in the decoder, `Advanced LE' method, which uses the advanced language tokens obtained by cross-attention with the vision features, showed better performance on each dataset. This suggests that enhancing the cross-modal alignment for language tokens helps to improve the comprehension for the meaning of language expression contexts. 
We compared our method with two methods: `Vanilla LE' method and `Advanced LE' method. As a key-value set in the decoder, `Vanilla LE' method uses the vanilla language encoder features, and `Advanced LE' method uses the advanced language tokens extracted by cross-attention with the vision features. As shown in Table \ref{tab:abltable1}, `Advanced LE' method showed better performance than `Vanilla LE' method on each dataset by improving the cross-modal alignment.

Different from these two ablation methods, our method uses the visual expression generated from the retrieved informative visual tokens. Compared to these ablation methods, our VE method showed remarkable gains by 4.19\% and 2.41\% oIoU on RefCOCOg, the most challenging dataset. From an attention alignment perspective, these gains suggest that our approach minimizes the modality gap between the query and key-value spaces, enhancing semantic consistency in attention.
These results also indicate that leveraging the visual expression as a key-value set provides the vision query features with semantic structural contexts of the informative regions more effectively, rather than projecting visual information into linguistic tokens (\textit{i.e}., advanced LE methods). 
Therefore, considering the informative parts of visual contexts in the attention mechanism of the segmentation decoder can improve the capability to guide the network towards the fine-grained region of interest on the referring segmentation task by enhancing attention coherence.

% \begin{figure}[t]
%     \centering 
% \begin{minipage}[t]{\linewidth}
% \centering
% %\hspace{0.2\textwidth}
% \includegraphics[width=\linewidth]{number_of_tokens.pdf}
% % \hspace{0.02\textwidth} 
% \vspace{-0.3cm}
% \subcaption{Performance by increasing the value of $r$}
% \end{minipage}
% \hfill%\hspace{0.02\columnwidth}
% \vspace{0.3cm}
% \begin{minipage}[t]{\linewidth}
% \centering 
% \includegraphics[width=\linewidth]{topk_iccv (1).pdf}
% % \hspace{0.02\textwidth}
% \vspace{-0.5cm}
% \subcaption{Segmentation results at different $r$}
% \end{minipage}
% \caption{Ablation study on the number of the retrieved informative visual tokens.}
% \label{topk}
% \end{figure}

\begin{table}[t]
\centering
\renewcommand{\arraystretch}{}
\captionsetup[table]{hypcap=false}
\captionof{table}{Ablation studies for the design of the visual expression generator module. Our default design is marked in \colorbox{lavender(web)}{\rule[0.2em]{0pt}{0.1em}  }.% \textcolor{green_color}{Drops} are relative to our default design.
}
\resizebox{\linewidth}{!}{\small
\begin{tabular}{C{0.8cm}C{1.3cm}C{1.3cm}C{1.3cm}C{1.3cm}C{1.3cm}C{1.3cm}}%C{1cm}C{1.5cm}C{2cm}C{2cm}C{2cm}C{2cm}C{2cm}C{2cm}
\toprule[1.3pt]
&\multicolumn{2}{c}{\textbf{Design}}&\multicolumn{2}{c}{\textbf{RefCOCO+  \textit{val}}}&\multicolumn{2}{c}{\textbf{{RefCOCOg \textit{val}}}}\\
\midrule
\multirow{4}{*}{(a)}&\multicolumn{1}{|c}{\textit{Step 1}}&\textit{Step 2}& {{mIoU}}& {{oIoU}}& {{mIoU}}& {{oIoU}}\\ 
\cline{2-7} \\[-0.95em]
%&\multicolumn{1}{|c}{{\xmark}}&{\xmark}&66.14 &64.16&65.06 &63.52\\
&\multicolumn{1}{|C{1.3cm}}{\xmark}&\cmark&66.67& 64.72& 66.31&64.04  \\
&\multicolumn{1}{|C{1.3cm}}{\cmark}&\xmark&66.29  & 64.54  &65.82 &63.79  \\
&\multicolumn{1}{|C{1.3cm}}{\cellcolor{lavender(web)}\cmark}&\cellcolor{lavender(web)}\cmark&\cellcolor{lavender(web)}\textbf{70.42}&\cellcolor{lavender(web)}\textbf{66.91}&\cellcolor{lavender(web)} \textbf{69.98}&\cellcolor{lavender(web)}\textbf{65.93}\\
\midrule[1.3pt]
\multirow{5}{*}{(b)}&\multicolumn{1}{|C{1.3cm}}{\textit{Global}} &\textit{Local}&{mIoU} & {oIoU}&{{mIoU}} & {{oIoU}} \\
\cline{2-7} \\[-0.95em]
&\multicolumn{1}{|C{1.3cm}}{\xmark}&\xmark& 66.67& 64.72& 66.31&64.04  \\
    &\multicolumn{1}{|C{1.3cm}}{\cmark}&\xmark&67.64 &65.15 & 67.60& 64.79\\
    &\multicolumn{1}{|C{1.3cm}}{\xmark}&\cmark&  68.05 	& 65.27 &67.33  &64.75  \\ 
    &\multicolumn{1}{|C{1.3cm}}{\cellcolor{lavender(web)}\cmark}&\cellcolor{lavender(web)}\cmark& \cellcolor{lavender(web)}\textbf{70.42}&\cellcolor{lavender(web)}\textbf{66.91} & \cellcolor{lavender(web)}\textbf{69.98}&\cellcolor{lavender(web)}\textbf{65.93}\\
    \bottomrule[1.3pt]
\end{tabular}}\label{tab:design}
\end{table}

\noindent
\textbf{Analysis on Design of Visual Expression Generator.}  
In Table \ref{tab:design} (a), the removal of Step 1 (\textit{i.e}., visual informative token retrieval) resulted in significant drops. This indicates that leveraging the retrieved informative parts of visual tokens is effective in guiding the network to the target region compared to using the total visual tokens, as most visual pixel tokens in the image are not in the region of interest.  
% These results indicate that it is effective to concentrating more on the informative tokens from the vision context that contains both the target-relevant information and the distracting non-target information. 
The removal of Step 2 (\textit{i.e}., visual context refinement) also resulted in notable performance degradation. This step enables the visual expression to capture the semantic visual contexts by mitigating the noise information and considering the visual relationship between each informative visual token.
%our refinement step  reducing the noise information and enhancing the visual relationship is effective  generate the visual expression that can provide target-relevant information as a key-value set. 
Therefore, these ablation results demonstrate that each step of our module is necessary to generate the informative visual expression.

% aggregating with refinement considering the retrieved visual token is more effective than simply aggregating the retrieved information by reducing the noisy information.
% This result highlights that adaptively capturing the semantic information from the retrieved informative token is more effective than simply aggregating the curated information for producing more semantic visual expression. 
% The removal of Step 3 resulted in a 1.82\% drop in oIoU on RefCOCO+. This indicates that each token of the visual expression acquires the visual context information regarding the target regions by considering the relationship between each visual token. These ablation studies demonstrate that each of the proposed phase is necessary to endow the visual expression.

In Table \ref{tab:design} (b), removing the use of the local linguistic cues showed  performance drops compared to our full model. Removing the use of the global linguistic cue also decreased performance.
% removing the use of local and global linguistic cues shows a 2.19\% and 1.89\% drop in oIoU compared to our full model on RefCOCO+ and RefCOCOg, respectively. 
Additionally, removing the use of both local-global linguistic cues, which exploits all vision tokes without the retrieval step, showed substantial drops in oIoU and mIoU.
% removing the use of the local linguistic cues showed a ~\% drop in oIoU compared to our full model on RefCOCO+. In addition, removing the use of the global linguistic cue showed a 2.03\% drop in oIoU on RefCOCO+. 
These results indicate that exploiting the linguistic context-relevant parts of visual contexts as a key-value set helps to effectively complement the limited target information of language inputs on referring image segmentation.
Furthermore, leveraging both global and local linguistic cues allows the visual expression tokens to consider both the comprehensive context and the attribute contexts for the enriched visual contexts of fine-grained target regions.

\begin{table}[t]\centering 
\renewcommand{\arraystretch}{}
\caption{Ablation for different encoder fusion types.}
\resizebox{\columnwidth}{!}
{
\small
\begin{tabular}{L{4cm}L{1.5cm}C{1.2cm}C{1.2cm}C{1.2cm}}
\toprule[1.3pt]
\multirow{2}{*}{\textbf{Encoder Feature Extraction}}  & \multirow{2}{*}{\textbf{Method}} & \multicolumn{3}{c}{\textbf{RefCOCO+}}\\ [-0.1em]
         \cmidrule(lr){3-5}   \\ [-1.2em]
&&\textbf{\textit{val}} & \textbf{\textit{test A}} & \textbf{\textit{test B}}\\
\midrule
\multirow{2}{*}{w/o Fusion} &w/o VE & 62.74 & 68.41 &55.82 \\
 &\cellcolor{lavender(web)}w/ VE &\cellcolor{lavender(web)}\textbf{65.15}&\cellcolor{lavender(web)}\textbf{71.17}&\cellcolor{lavender(web)}\textbf{58.34}\\
\midrule
\multirow{2}{*}{Late Fusion} &w/o VE&63.88&69.65&57.07\\
&\cellcolor{lavender(web)}w/ VE  &\cellcolor{lavender(web)}\textbf{66.74} & \cellcolor{lavender(web)}\textbf{72.21}&\cellcolor{lavender(web)}\textbf{59.73}\\
\midrule
% \multirow{2}{*}{Unidirectional Early Fusion}& w/o VE &63.97&69.51&56.92\\
% &\cellcolor{lavender(web)}w/ VE (Ours) &\cellcolor{lavender(web)}\textbf{66.70}& \cellcolor{lavender(web)}\textbf{72.08}& \cellcolor{lavender(web)}\textbf{59.85}\\
% \midrule
\multirow{2}{*}{Early Fusion}&  w/o VE & 64.16& 69.86&57.39\\
&\cellcolor{lavender(web)}w/ VE&\cellcolor{lavender(web)}\textbf{66.91} & \cellcolor{lavender(web)}\textbf{72.44} & \cellcolor{lavender(web)}\textbf{60.15}\\
\bottomrule[1.3pt]
\end{tabular}
}
\label{tab:featureextract}
\end{table}

\begin{figure*}[t]
    \centering 
\begin{minipage}[t]{\columnwidth}
\centering
%\hspace{0.2\textwidth}
\raisebox{0cm}{
\includegraphics[width=\columnwidth]{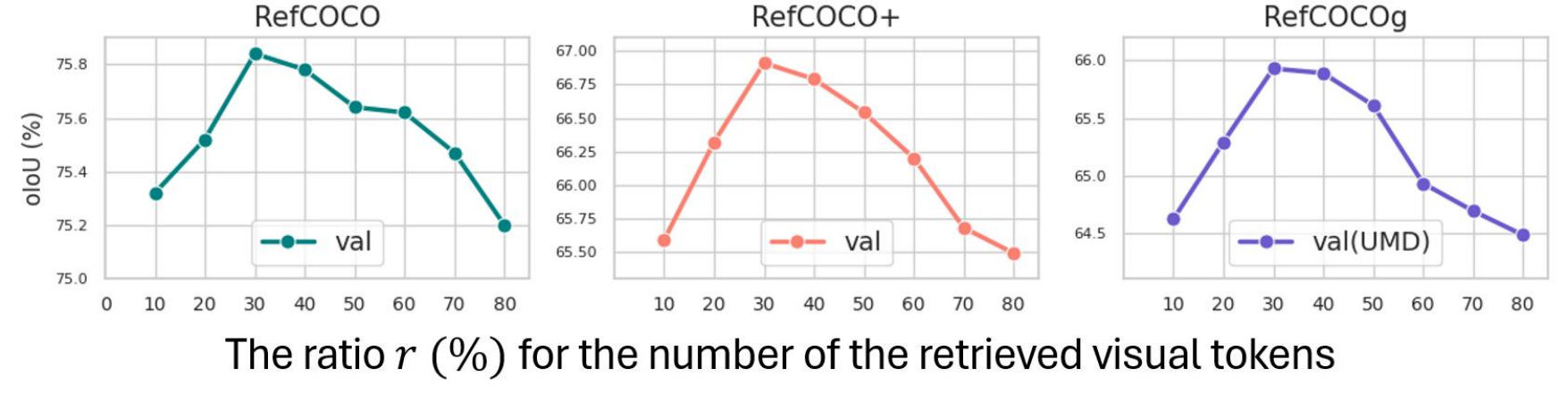}}
% \hspace{0.02\textwidth}
% \vspace{-0.5cm}
\subcaption{Performance by increasing the value of $r$}
\end{minipage}\hfill%\hspace{0.02\columnwidth}
\begin{minipage}[t]{\columnwidth}
\centering 
\includegraphics[width=\columnwidth]{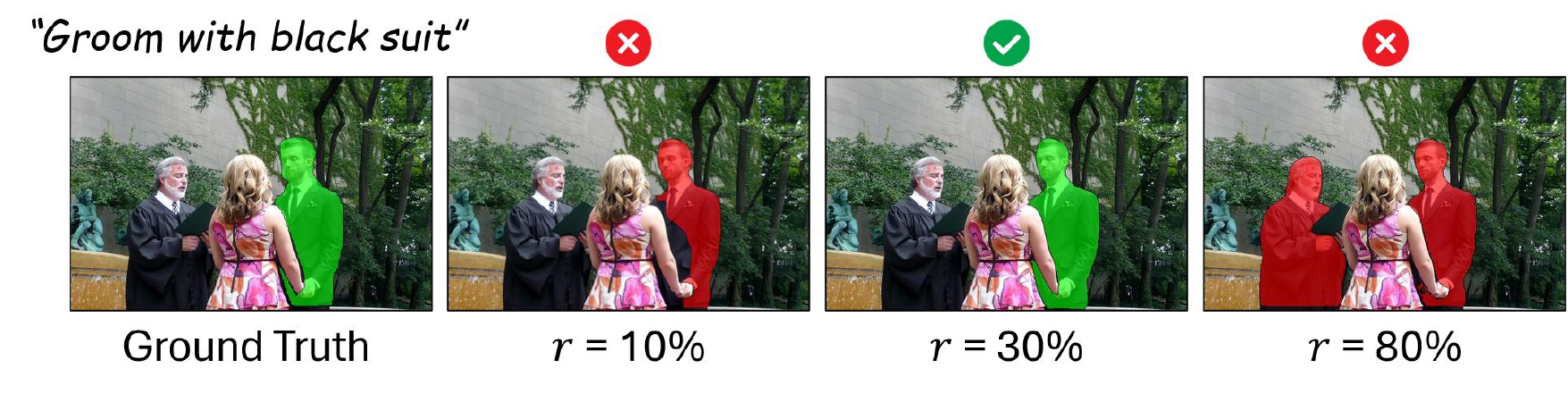}
% \vspace{-0.5cm}% \hspace{0.02\textwidth}
\subcaption{Segmentation results at different $r$}
\end{minipage}
\caption{Ablation study on the number of the retrieved informative visual tokens.}
\label{topk}
\end{figure*}

\begin{table}[t]
\centering
\renewcommand{\arraystretch}{}
\caption{Additional ablation on supervision by the contrastive loss. Our default design is marked in \colorbox{lavender(web)}{\rule[0.2em]{0pt}{0.1em}  }.}
\resizebox{\linewidth}{!}{\small
\begin{tabular}{lccc}%
\toprule[1.3pt]
{\textbf{{Method}}}  & {\textbf{RefCOCO+ \textit{val}}} & {\textbf{RefCOCO+ \textit{test A}}}&{\textbf{RefCOCO+ \textit{test B}}}\\
\midrule
w/o Supervision &65.93& 71.50&59.26\\
Direct thresholding loss & 66.52 & 71.93 & 59.60\\
\cellcolor{lavender(web)}Pixel contrastive loss&\cellcolor{lavender(web)}\textbf{66.91} & \cellcolor{lavender(web)}\textbf{72.44}  & \cellcolor{lavender(web)}\textbf{60.15} \\

% \cellcolor{lavender(web)}w/ contrastive loss&\cellcolor{lavender(web)}\textbf{68.63} &\cellcolor{lavender(web)}\textbf{66.70}\\
    \bottomrule[1.3pt]
\end{tabular}}
\label{tab:design_appendix}
\end{table}

\begin{table}[t]
    \centering
\renewcommand{\arraystretch}{}
\caption{Ablation study on the use of the article tokens at the process of retrieving informative visual contexts. Our default design is marked in \colorbox{lavender(web)}{\rule[0.2em]{0pt}{0.1em}  }.}
\resizebox{\linewidth}{!}{\small
\begin{tabular}{L{1.6cm}C{1.1cm}C{1.1cm}C{1.1cm}C{1.1cm}C{1.1cm}C{1.1cm}}
\toprule[1.3pt]
        \multirow{2}{*}{\textbf{{Method}}}  & \multicolumn{2}{c}{\textbf{RefCOCO \textit{val}}}& \multicolumn{2}{c}{\textbf{RefCOCO+ \textit{val}}} & \multicolumn{2}{c}{\textbf{RefCOCOg \textit{val}}} \\[-0.1em]
         \cmidrule(lr){2-3} \cmidrule(lr){4-5} \cmidrule(lr){6-7}  \\ [-1.2em]
        &{mIoU} & {oIoU} &{mIoU} & {oIoU} &{mIoU} & {oIoU}  \\
        \midrule
         {w/o articles}& 77.80 &75.11  & 69.35 	&66.32 &	67.74 &	64.99  \\
         \cellcolor{lavender(web)}{All words} &\cellcolor{lavender(web)}\textbf{78.68}	&\cellcolor{lavender(web)}\textbf{75.84}&	\cellcolor{lavender(web)}\textbf{70.42}&	\cellcolor{lavender(web)}\textbf{66.91}&	\cellcolor{lavender(web)}\textbf{69.98}&	\cellcolor{lavender(web)}\textbf{65.93}\\
         \bottomrule[1.3pt]
    \end{tabular}}
    \label{tab:articles}
\end{table}

\noindent
\textbf{Encoder-type Agnostic.} We experimented with different encoder types in Table \ref{tab:featureextract} to verify the versatility of our method and to analyze the impact of vision-language encoder alignment. Our method with early fusion showed higher performance than other types. Notably, across all encoder types, the introduction of visual expression consistently improves performance, including the setting without any explicit fusion in the encoder. This result demonstrates that our visual expression approach does not rely on a specific encoder design and can functions as an effective and complementary component regardless of the fusion strategy. Therefore, VIPA is encoder-type agnostic and can robustly enhance segmentation performance across diverse architectural choices.

\noindent
\textbf{Number of Retrieved Tokens.} We analyzed the ratio $r$ for the number of the retrieved visual tokens. Compared to the $r$ values of 10 and 80, the $r$ of 30 showed higher oIoU in Fig. \ref{topk} (a). In Fig. \ref{topk} (b), the $r$ of 30 segmented more clearly, while the $r$ of 10 missed some part of the target regions and the $r$ of 80 even segmented other object regions. The smaller number of $r$ resulted in a lack of information, where the informative visual tokens cannot be sufficiently exploited. In contrast, the larger number of $r$ resulted in higher proportion of noise information and distracting the informativeness of the visual expression. Thus, the optimal $r$ can exploit semantic visual information and filter out noise components to improve the robustness of the guidance capacity.

\noindent
\textbf{Ablation for Supervision by Contrastive Loss.} 
In Table \ref{tab:design_appendix}, we experimented on supervising the relevance score map by the pixel contrastive loss. The contrastive loss is more effective to monitor the retrieval of informative visual tokens, compared to the direct thresholding loss. In addition, the contrastive loss helps to monitor the curation of the informative tokens associated with the correct target region and to prevent the high relevance scores between the linguistic features and incorrect regions. 

\noindent
\textbf{Ablation for Using Article of Language.} 
we experimented the ablation on the use of the article tokens such as “the”, “a” and “an”, which are meaningless words in the input sentence, in the process of the visual informative token retrieval. As shown in Table \ref{tab:articles}, compared to using all word tokens, ‘w/o article’ resulted in 0.73\%, 0.59\% and 0.94\% drops in oIoU on each dataset, respectively. These results indicate that the article tokens do not carry the noise information, and using all word tokens as linguistic cues are more effective at collecting the informative visual tokens. Since the relations of each word are considered during encoding the language input to capture the contextual information for the target object description, each language token is encoded with semantic representations.

\begin{figure*}[t]
\centering
\includegraphics[width=\linewidth]{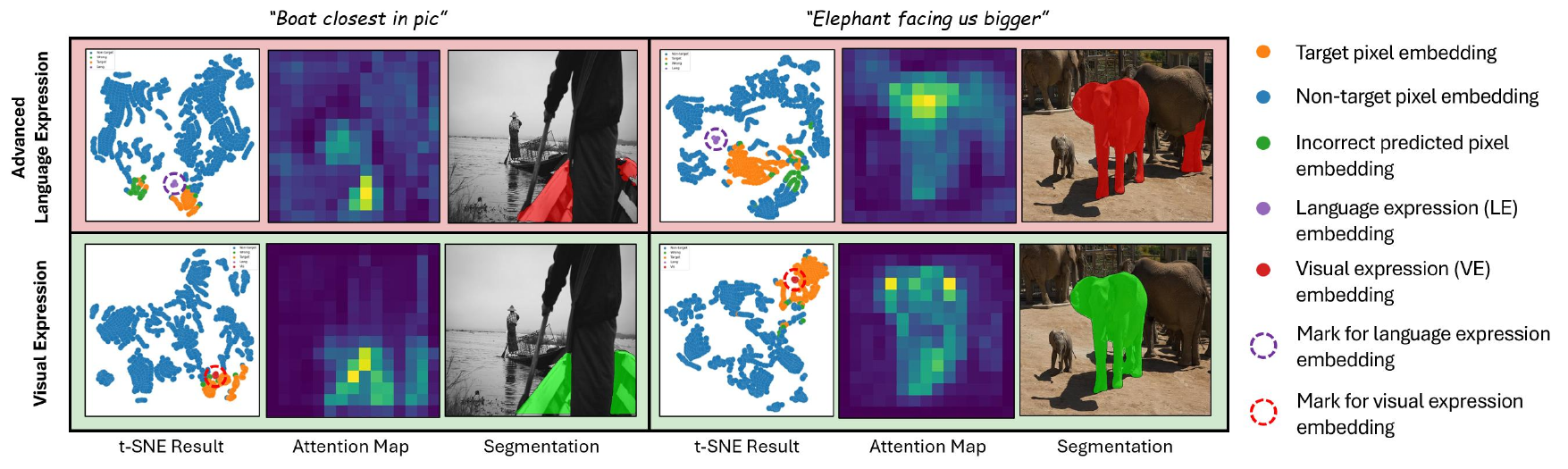} 
\caption{t-SNE visualizations and qualitative comparisons between the ablation method (\textit{i.e}., advanced language expression) and our method (\textit{i.e}., visual expression).}
\label{tsne}
\end{figure*}

\begin{figure*}[t]
\includegraphics[width=\linewidth]{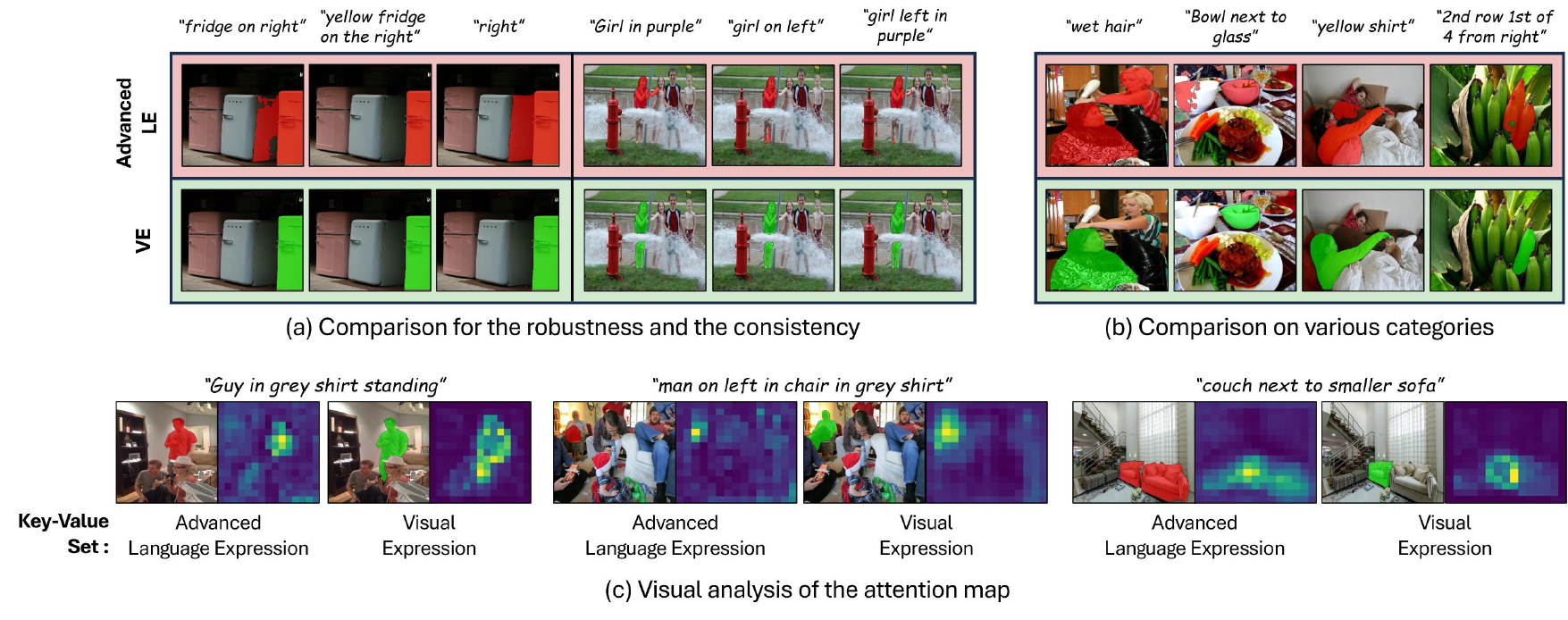} 
\caption{(a-b) Visualized comparison with the ablation method (\textit{i.e}., advanced Language Expression (LE)) and our method (\textit{i.e}., Visual Expression (VE)). (c) Visual analysis of the attention map between vision query features and visual expression key-value set, and the attention map between vision query features and advanced language expression key-value set.}
\label{ablation}
\end{figure*}

\subsection{Qualitative Results}
\label{qual_section}

\noindent
\textbf{t-SNE Visualizations.}
In Fig. \ref{tsne}, we presented t-SNE visualizations and qualitative comparisons between our visual expression method and the ablated method that employs advanced language expression. The t-SNE exhibits the distribution of pixel tokens and visual and linguistic expression tokens in the embedding space. As shown in the t-SNE results, visual expression embeddings are closer to the target pixels, whereas the linguistic expression embeddings of the ablated model exhibit weaker alignment with target pixel embeddings. 

Consistent with these observations, the qualitative results further demonstrate the advantage of visual expression. The ablated model either segmented only partial target regions (top row) or incorrectly attended to non-target regions (bottom row). In contrast, our method produces more accurate and complete segmentations by effectively incorporating semantically informative visual contexts. These results indicate that our visual informative part attention approach enhances semantic consistency in attention, leading to more robust segmentation performance.

\noindent
\textbf{Visual Comparisons with Ablation Method.} We compared with the ablation method (\textit{i.e}., advanced language expression) on different language expressions describing the same target object and on various target object categories in Fig. \ref{ablation} (a) and (b), respectively. The ablation method inconsistently predicted the target regions and showed incorrect segmentation.  In contrast, our VE method consistently predicted accurate regions for both large target regions and small target regions. These results indicate that our method enhances the adaptability to diverse language and image inputs.

\begin{figure*}[t]
\centering
\includegraphics[width=\linewidth]{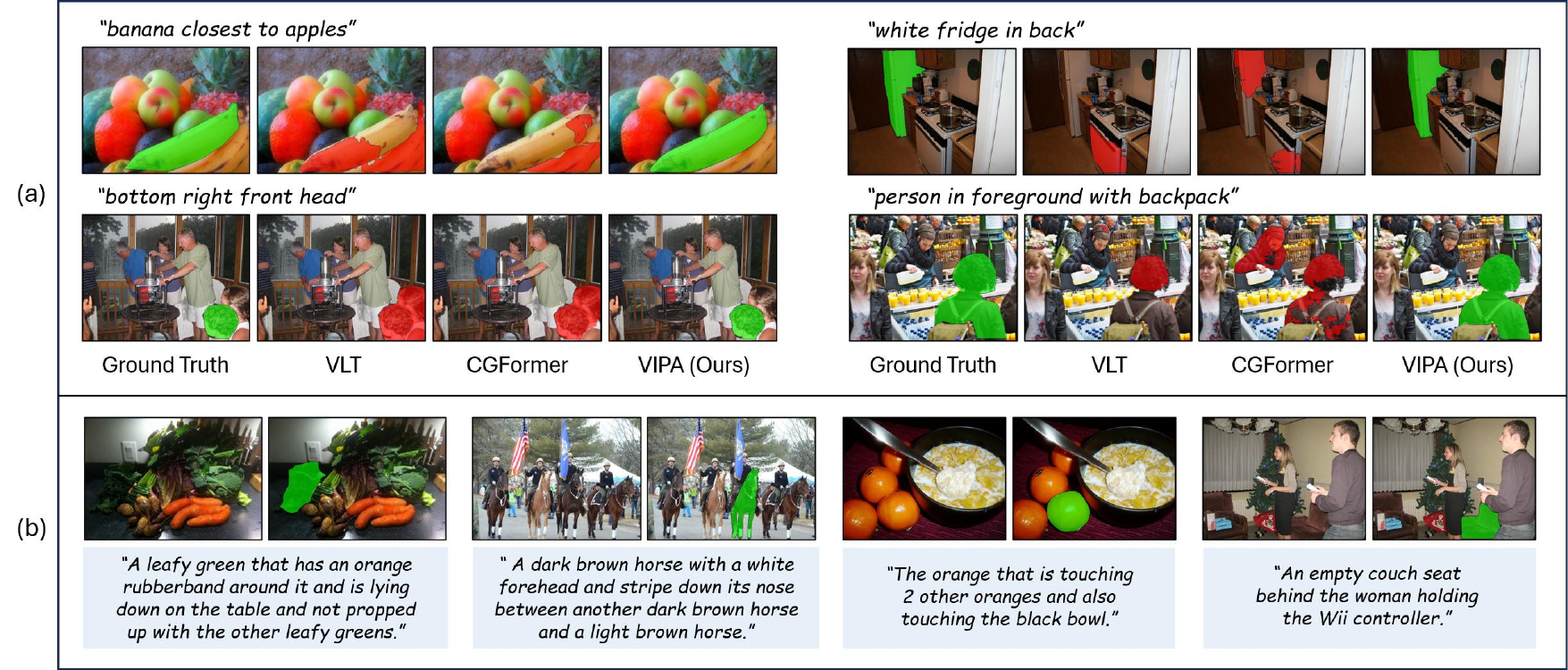} 
\caption{(a) Qualitative comparison with Transformer-based RIS methods. (b) Our results on long and difficult language inputs.}
\label{others}
\end{figure*}

\noindent
\textbf{Visual analysis of Attention maps.} We presented additional visual analysis of the attention maps in Fig. \ref{ablation} (c): one between  the vision query (Q) features and the key-value (KV) set of the visual expression, and another between the vision query (Q) features and the key-value (KV) set of the advanced language expression.
The visual expression robustly guides the network's attention to the regions of interest, whereas the language expression fails to focus on the real target regions or highlights even wide non-target regions. These qualitative results support that our VIPA approach theoretically improves attention coherence and semantic consistency. 

Concretely, the informative parts of the visual contexts, \textit{visual expression}, provide KV representations that are aligned in the visual feature space of the query Q, yielding lower modality projection entropy than the language-based KVs.
Therefore, this visual analysis demonstrates that our approach more effectively exploits the structural information for fine-grained segmentation, compared to the existing approach using visual-aware linguistic tokens.

\noindent
\textbf{Qualitative Comparisons with Transformer-based RIS methods.}
In Fig. \ref{others} (a), we compared with previous methods using the advanced linguistic tokens as a key-value set. Ours segmented more clearly, whereas others incorrectly predicted and uncertainly segmented the regions. These indicate that our approach is more effective in improving visual understanding of the fine-grained target regions. In Fig. \ref{others} (b), we visualized our results on longer and complex language inputs with the complicated images. These results indicate that \textsc{\method} is robust in challenging scenarios. 
All of these visual comparisons demonstrate that the visual expression can effectively serve as an advanced information provider in the attention mechanism of referring image segmentation. More various qualitative comparisons are provided in supplementary materials.

\section{Conclusion}
We propose a novel Visual Informative Part Attention (VIPA) framework to more effectively exploit visual contexts for referring image segmentation. Our VIPA is the first study to leverage \textit{visual expression}, generated from the informative parts of visual contexts, as a key-value set for the vision query feature in the attention mechanism of the referring segmentation decoder. The visual expression provides the key-value representations that are aligned in the visual feature space of the query.
From an attention alignment standpoint, this design aligns the query and the key-value set within the visual modality space, reducing high-variance cross-modal projection and enhancing semantic consistency.
Therefore, our approach robustly guides the network's attention to fine-grained target regions by providing structural and semantic visual information. 
Extensive comparisons and ablations demonstrate the effectiveness of exploiting retrieved informative visual parts for Transformer-based referring image segmentation.

% \section*{Acknowledgments}
% This work was supported by Samsung Electronics Co., Ltd. (IO251218-14799-01). 
% This research was also supported by the MSIT (Ministry of Science and ICT), Korea, 
% under the ITRC (Information Technology Research Center) support program 
% (IITP-2025-RS-2023-00260091) supervised by the IITP 
% (Institute for Information \& Communications Technology Planning \& Evaluation).

% \section{References Section}
% You can use a bibliography generated by BibTeX as a .bbl file.
%  BibTeX documentation can be easily obtained at:
%  http://mirror.ctan.org/biblio/bibtex/contrib/doc/
%  The IEEEtran BibTeX style support page is:
%  http://www.michaelshell.org/tex/ieeetran/bibtex/
 
% You can manually copy in the resultant .bbl file and set second argument of $\backslash${\tt{begin}} to the number of references
%  (used to reserve space for the reference number labels box).

\bibliographystyle{IEEEtran}

\bibliography{IEEEabrv, reference}
% \end{thebibliography}

\newpage

\begin{IEEEbiography}[{\noindent\includegraphics[width=1in,height=1.25in,clip,keepaspectratio]{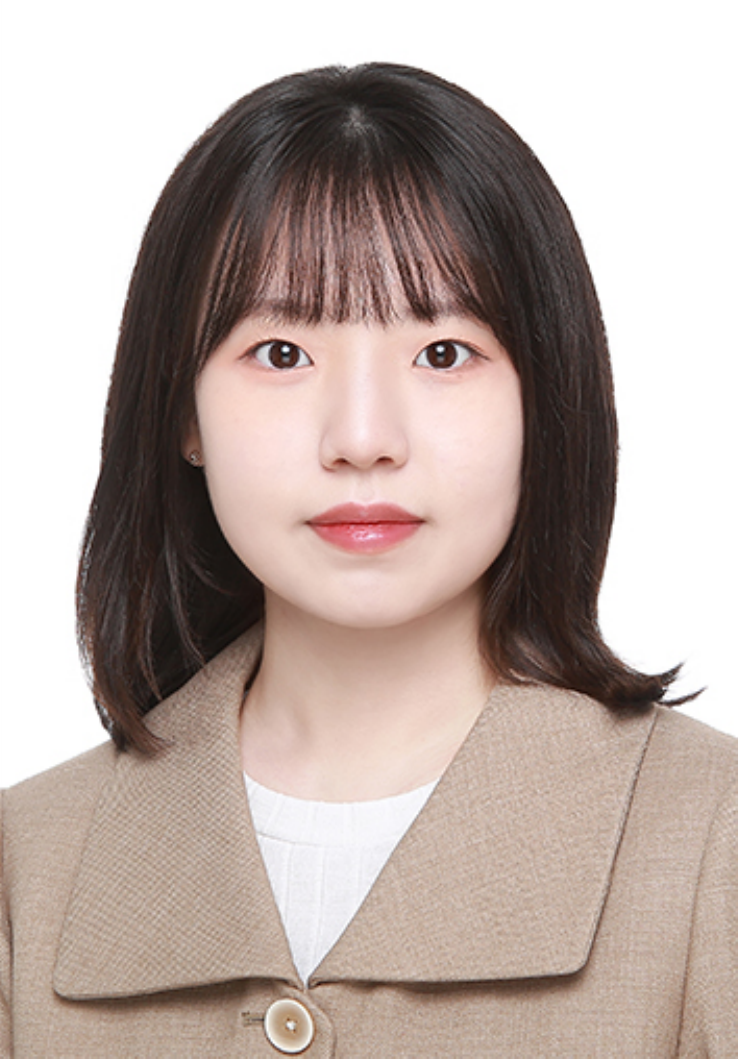}}]{Yubin Cho} received the B.S. degree with double major in Mechanical Engineering and Artificial Intelligence from Sogang University, South Korea, in 2022, and the M.S. degree with the School of Artificial Intelligence, Sogang University, in 2024. She is currently with the AI Lab of CTO division, LG Electronics.
Her current research interests include computer vision, multi-modal learning and embodied AI.
\end{IEEEbiography}

{\vskip -1\baselineskip plus -1fil}

\begin{IEEEbiography}[{\noindent\includegraphics[width=1in,height=1.25in,clip,keepaspectratio]{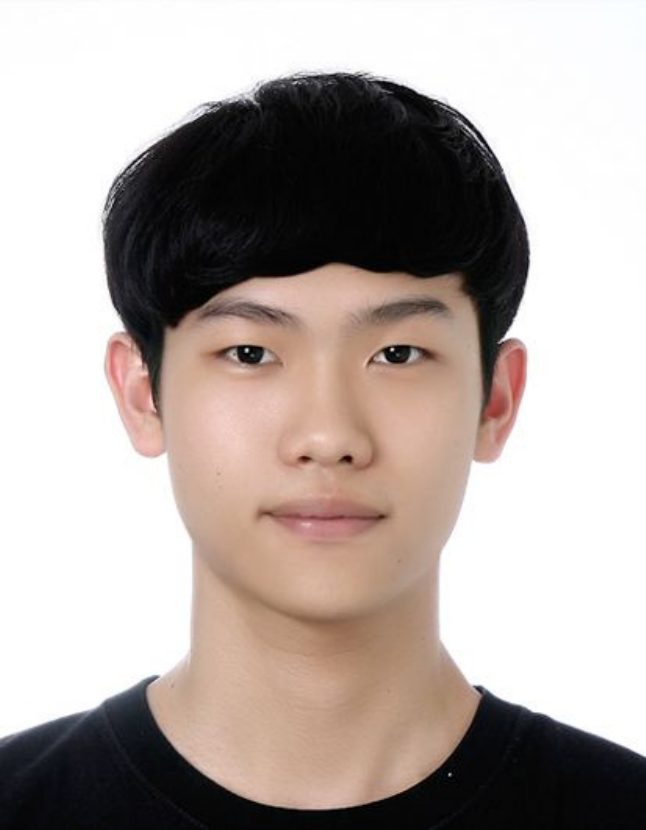}}]{Hyunwoo Yu} received the B.S. degree in Physics from Kangwon University, South Korea, in 2021. He is currently working toward the Ph.D. degree with the School of Electronic Engineering, Sogang University, South Korea. His current research interests include computer vision, multi-modal learning and pixel-level scene understanding.
\end{IEEEbiography}

{\vskip -1\baselineskip plus -1fil}

\begin{IEEEbiography}[{\noindent\includegraphics[width=1in,height=1.25in,clip,keepaspectratio]{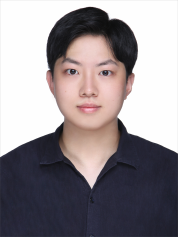}}]{Kyeongbo Kong} 
received the B.S. degree in electronics engineering from Sogang University, Seoul, South Korea, in 2015, and the M.S. and Ph.D. degrees in electrical engineering from the Pohang University of Science and Technology (POSTECH), Pohang, South Korea, in 2017 and 2020, respectively. From 2020 to 2021, he was worked as a Postdoctoral Fellow with the Department of Electrical Engineering, POSTECH, Pohang, South Korea. From 2021 to 2023, he was an Assistant Professor of Media School at Pukyong National University, Busan. He is currently an Associate Professor of Electrical and Electronics Engineering at Pusan National University. His current research interests include image processing, computer vision, machine learning, and deep learning.
\end{IEEEbiography}

{\vskip -1\baselineskip plus -1fil}

\begin{IEEEbiography}[{\noindent\includegraphics[width=1in,height=1.25in,clip,keepaspectratio]{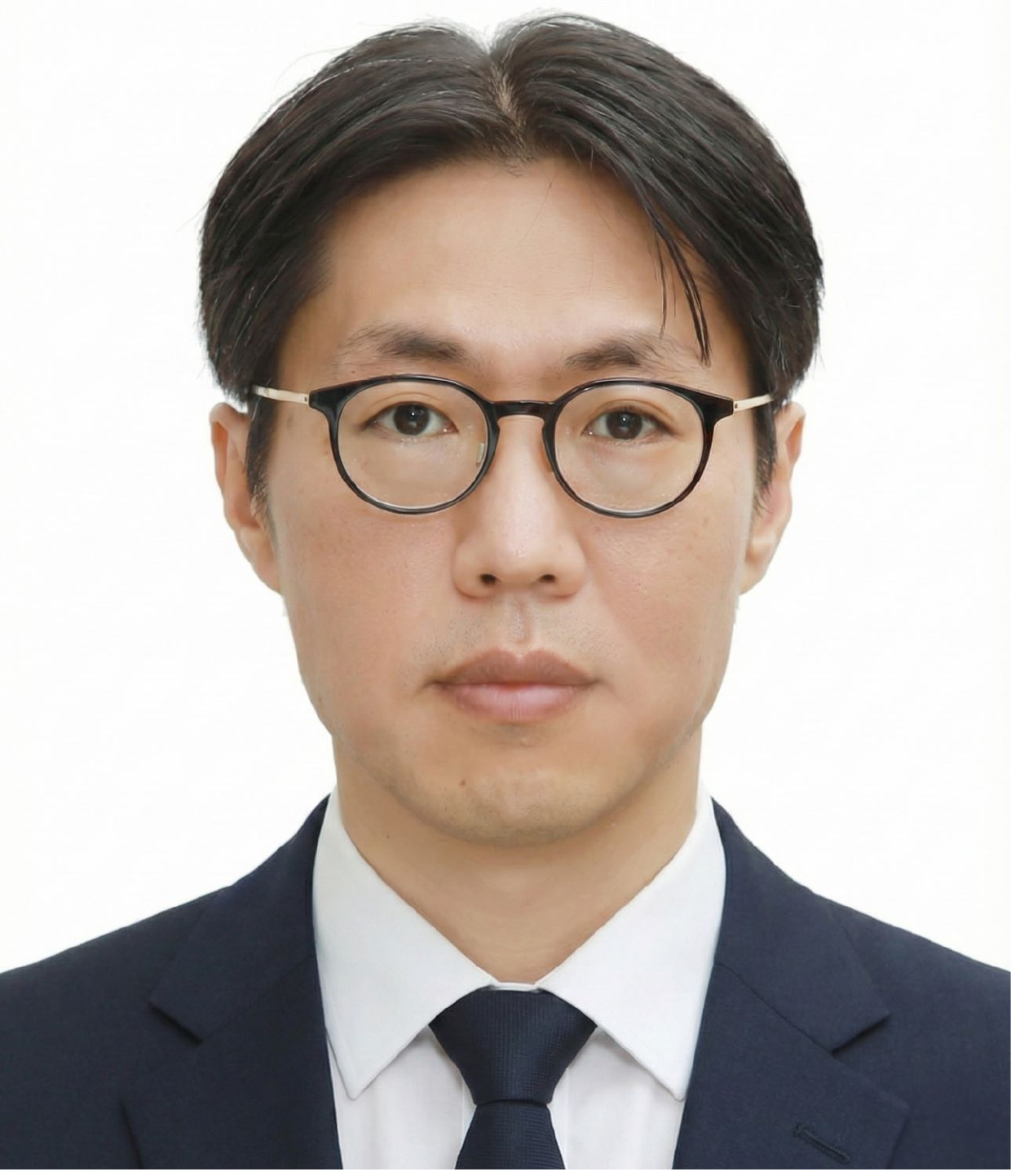}}]{Kyomin Sohn} 
is a Samsung Master (VP of Technology) in Samsung Electronics and he is responsible for future architecture and circuit technology of DRAM. He received the B.S. and M.S. degrees in Electrical Engineering in 1994 and 1996, respectively, from Yonsei University, Seoul. From 1996 to 2003, he was with Samsung Electronics, Korea, involved in SRAM Design Team. He designed various kinds of high-speed SRAMs. He received the Ph.D. degree in EECS in 2007 from KAIST, Korea. He rejoined Samsung Electronics in 2007, where he has been involved in DRAM Design Team. He led the development of HBM2 DRAMs and HBM-PIM. His interests include 3D-DRAM, reliable memory design, and processing-in-memory (PIM). In addition, he has currently served as a Technical Program Committee member of Symposium on VLSI Circuits since 2012.
\end{IEEEbiography}

{\vskip -1\baselineskip plus -1fil}

\begin{IEEEbiography}[{\noindent\includegraphics[width=1in,height=1.25in,clip,keepaspectratio]{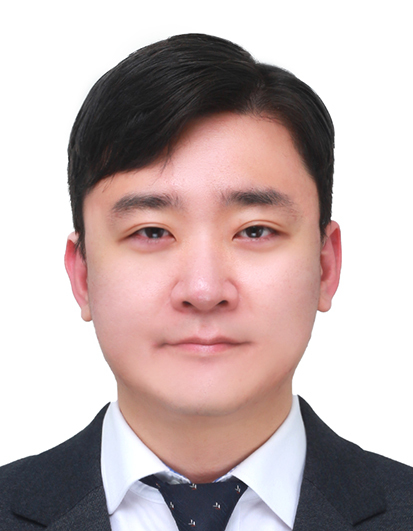}}]{Bongjoon Hyun}  received the B.S. degree in mechanical engineering and the M.S. degree in creative IT engineering from Pohang University of Science and Technology (POSTECH), Pohang, South Korea, in 2016 and 2018, respectively, and the Ph.D. degree in electrical engineering from the Korea Advanced Institute of Science and Technology (KAIST), Daejeon, South Korea, in 2025. He is currently a Staff Engineer at Samsung Electronics. His research interests include Processing-in-Memory (PIM) architectures for AI systems, with a specific focus on Large Language Model (LLM) serving systems.
\end{IEEEbiography}

{\vskip -1\baselineskip plus -1fil}

\begin{IEEEbiography}[{\noindent\includegraphics[width=1in,height=1.25in,clip,keepaspectratio]{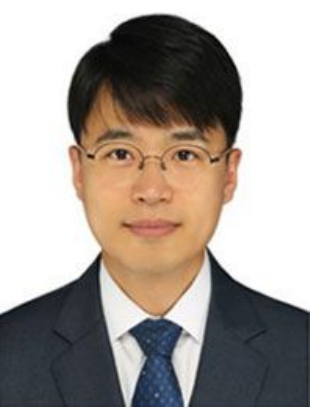}}]{Suk-ju Kang}
(Member, IEEE) received the B.S. degree in electronic engineering from Sogang University, Seoul, South Korea, in 2006, and the Ph.D. degree in electrical and computer engineering from the Pohang University of Science and Technology, Pohang, South Korea, in 2011. From 2011 to 2012, he was a Senior Researcher with LG Display Co., Ltd., Seoul, where he was a Project Leader for resolution enhancement and multiview 3-D system projects. From 2012 to 2015, he was an Assistant Professor of Electrical Engineering with Dong-A University, Busan, South Korea. He is currently a Professor of Electronic Engineering with Sogang University, Seoul. His current research interests include  computer vision, image analysis and enhancement, video processing, multimedia signal processing, digital system design, and deep learning. He was a recipient of the IEIE/IEEE Joint Award for Young IT Engineer of the Year in 2019 and the Merck Young Scientist Award in 2022. He served as an Associate Editor for IEEE Transactions on Circuits and Systems for Video Technology from 2023. 
\end{IEEEbiography}

 

% \end{IEEEbiography}



\vfill

\clearpage

\appendix

\section{Additional Qualitative Results}\label{app_a}
\begin{figure*}
    \centering
    \includegraphics[width=0.65\linewidth]{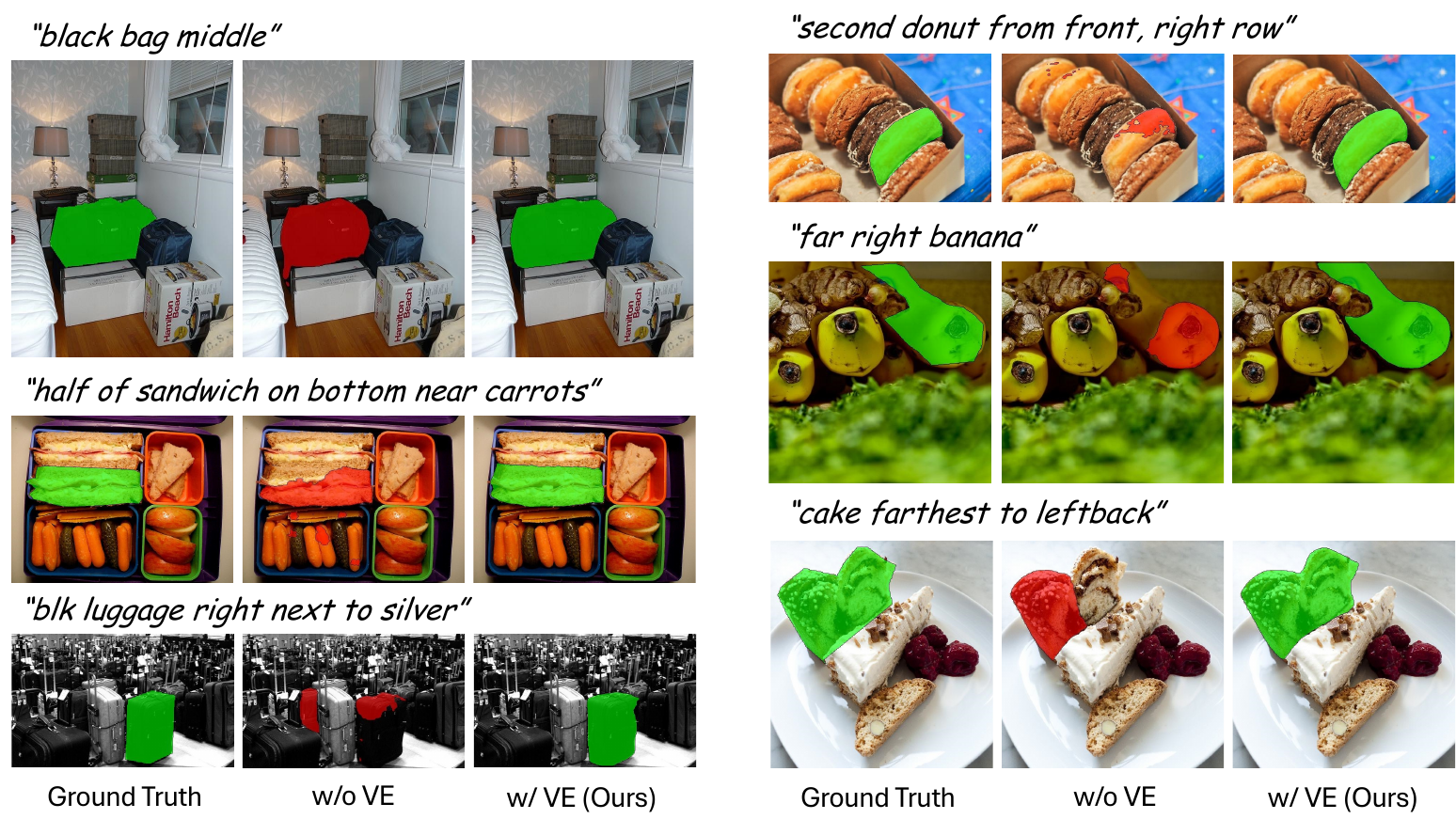} \vspace{-0.2cm}
    \caption{Additional visualization comparison of our method and the ablated method on various target object categories, where the ablation model  misses the target regions.} 
    \label{fig:miss_target}
\end{figure*}

\begin{figure*}
    \centering
    \includegraphics[width=0.65\linewidth]{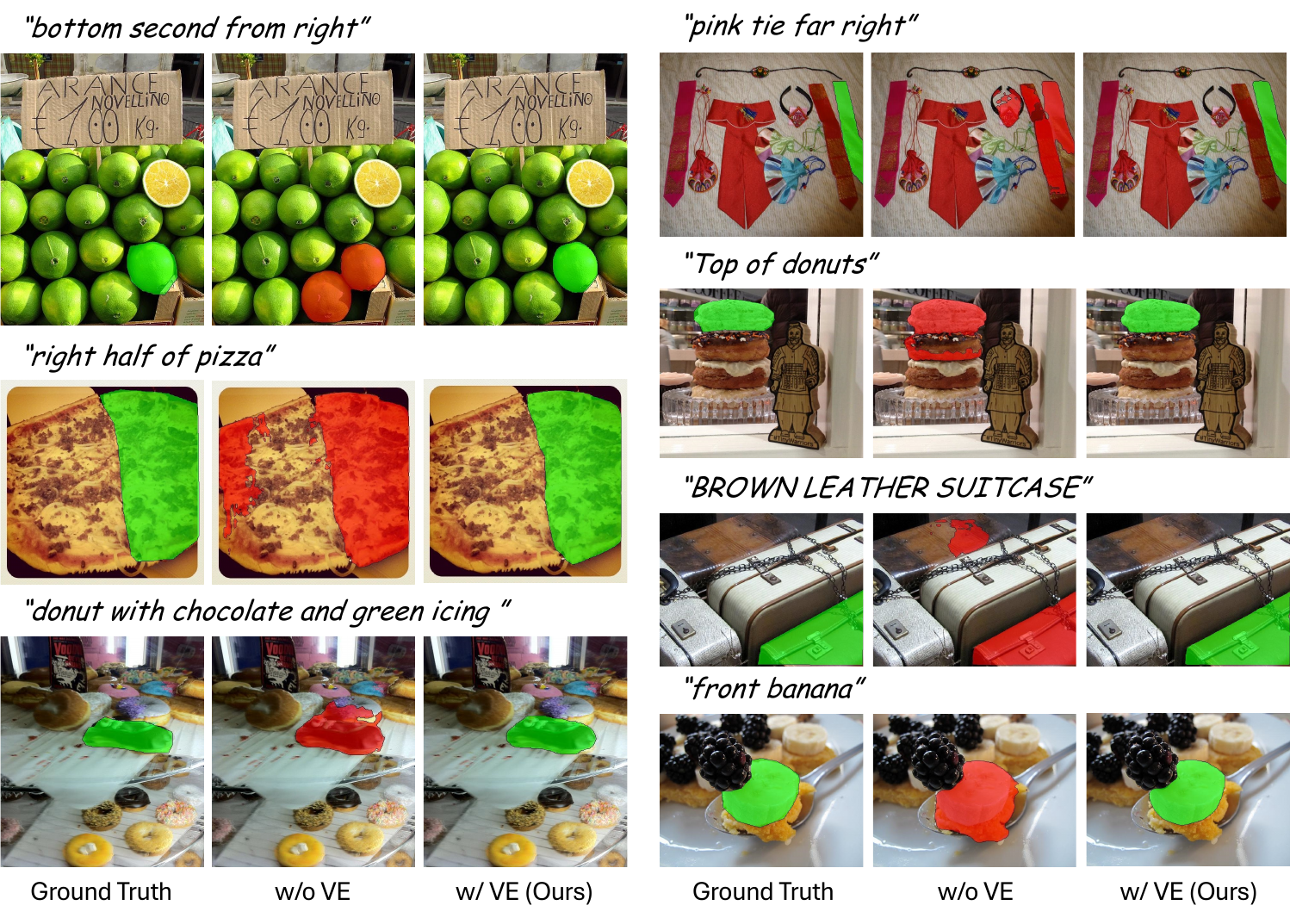} \vspace{-0.2cm}
    \caption{Additional visualization comparison of our method and the ablated method on various target object categories, where the ablation model segments even non-target regions.}
    \label{fig:non_target}
\end{figure*}

\begin{figure*}
    \centering
    \includegraphics[width=\linewidth]{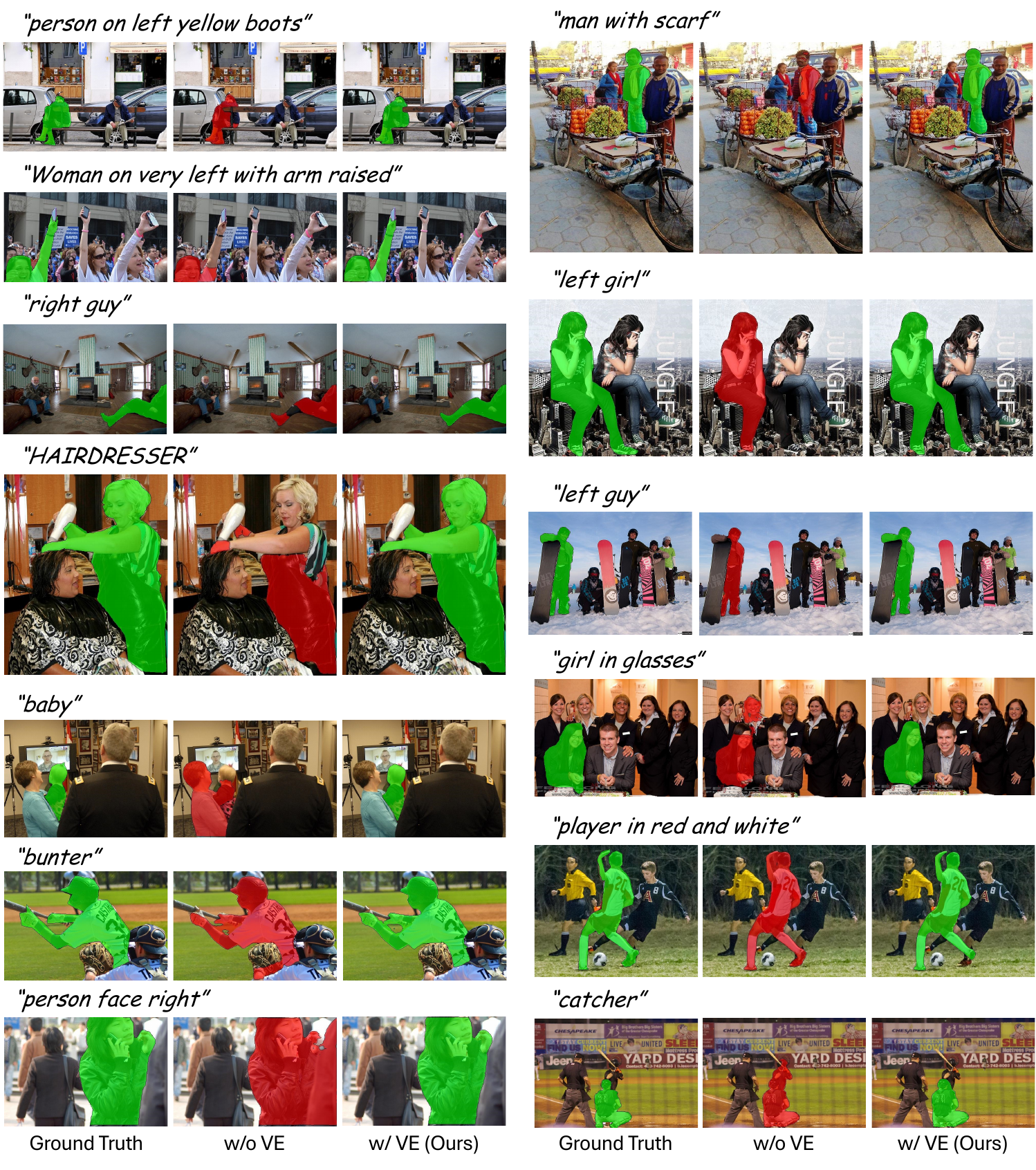}
    \caption{Additional visualization comparison of our method and the ablated method on the target object of the person, where the ablation model fails to capture the target regions.}
    \label{fig:person}
\end{figure*}

\begin{figure*}
    \centering
    \includegraphics[width=\linewidth]{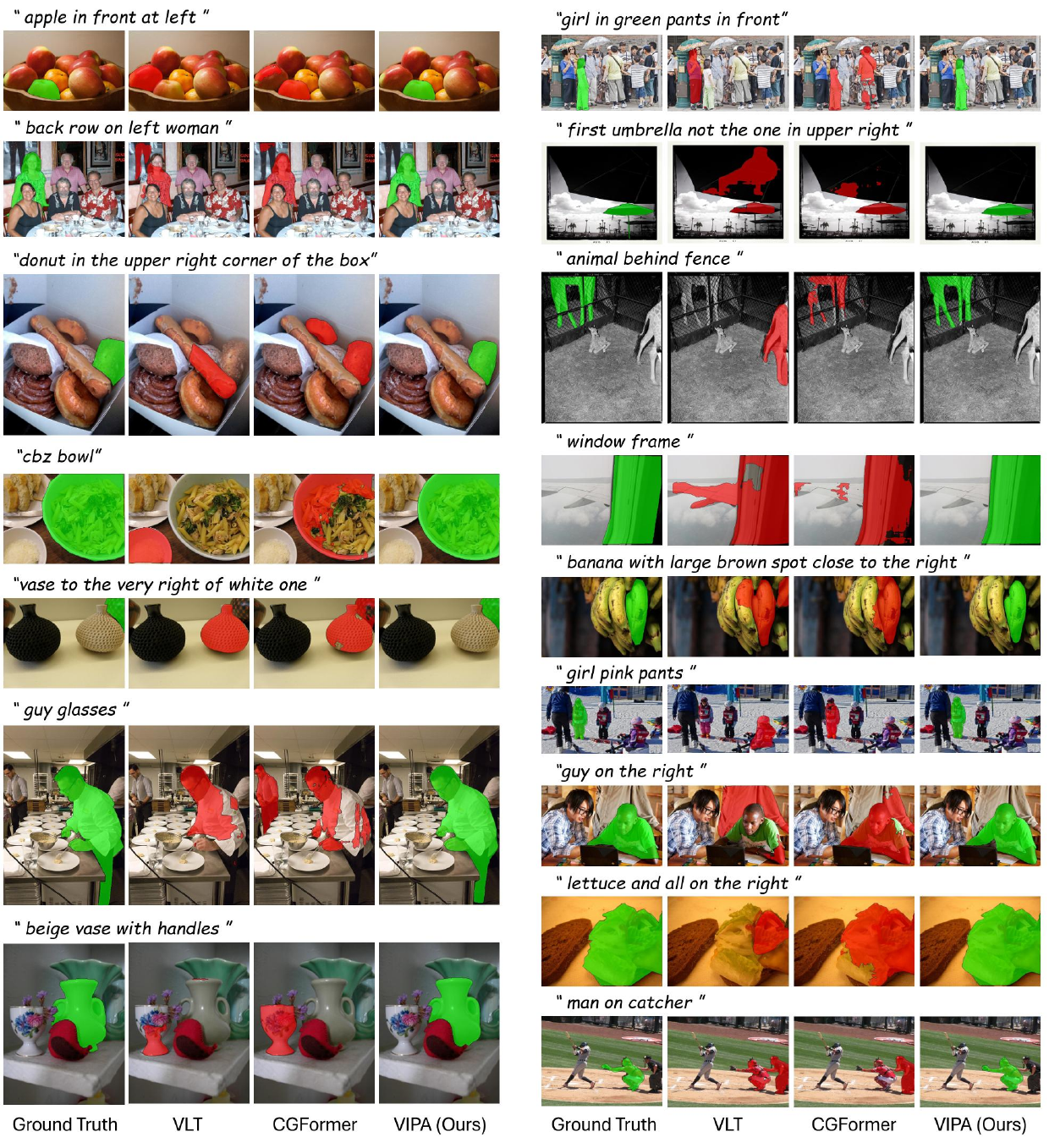}
    \caption{Additional qualitative comparison with previous referring image segmentation models.}
    \label{fig:ablation_visual}
\end{figure*}

In Figs. \ref{fig:miss_target}, \ref{fig:non_target} and \ref{fig:person}, we visualized the additional comparisons of our method and the ablated method, which uses the advanced language expression as a key-value set in the Transformer-based segmentation decoder, on various scenarios. The ablated model segment non-target region or misses the target region, whereas our visual expression method effectively segmented the target region. 
In Fig. \ref{fig:ablation_visual}, we compared our method with previous Transformer-based RIS approaches that exploit advanced language tokens as a key-value set on diverse image and language inputs. Our approach showed clearer segmentation compared to other methods. These results indicate that our method enhances visual understanding of fine-grained target regions more effectively.

% \appendices

% \section{Additional Qualitative Results}\label{app_a}
% In Figs. \ref{fig:miss_target}, \ref{fig:non_target} and \ref{fig:person}, we visualized the additional comparisons of our method and the ablated method, which uses the advanced language expression as a key-value set in the Transformer-based segmentation decoder, on various scenarios. The ablated model segment non-target region or misses the target region, whereas our visual expression method effectively segmented the target region. 
% In Fig. \ref{fig:ablation_visual}, we compared our method with previous Transformer-based RIS approaches that exploit advanced language tokens as a key-value set on diverse image and language inputs. Our approach showed clearer segmentation compared to other methods. These results indicate that our method enhances visual understanding of fine-grained target regions more effectively.

\end{document}